\documentclass{article}

\usepackage[preprint]{corl_2026} 
\usepackage{amsmath}
\usepackage{amssymb}
\usepackage{booktabs}
\usepackage{multirow}
\usepackage{graphicx}
\usepackage{subcaption}
\usepackage{float}
\usepackage{xspace}
\usepackage{tikz}
\usetikzlibrary{arrows.meta,positioning,calc,decorations.pathreplacing}

\newcommand{\hugsim}{HUGSim\xspace}
\newcommand{\nuscenes}{nuScenes\xspace}
\newcommand{\nuplan}{nuPlan\xspace}

\newcommand{\hdscore}{HD-Score\xspace}

\newcommand{\terrazero}{TerraZero\xspace}

\newcommand{\uniad}{UniAD\xspace}
\newcommand{\vad}{VAD\xspace}
\newcommand{\ltf}{LTF\xspace}
\newcommand{\eco}{ECO\xspace}

\newcommand{\svd}{SVD\xspace}

\newcommand{\ttc}{TTC\xspace}
\newcommand{\comscore}{COM\xspace}

\newcommand{\rc}{\ensuremath{R_c}\xspace}

\providecommand{\vs}{vs.\xspace}

\providecommand{\defacto}{\emph{de facto}\xspace}        
\providecommand{\perse}{\emph{per se}\xspace}            
\providecommand{\sinequanon}{\emph{sine qua non}\xspace} 
\providecommand{\tabularasa}{\emph{tabula rasa}\xspace}  

\title{TerraTransfer: Learning End-to-End Driving Policies\protect\\
Without Expert Demonstrations}

\author{
  Zikang Xiong\textsuperscript{1}, \;
  Weixin Li\textsuperscript{1}, \;
  Zhouchonghao Wu\textsuperscript{1}, \;
  Akshay Rangesh\textsuperscript{1}, \;
  Saarth Bonde\textsuperscript{1}, \\
  \bfseries
  Grantland Hall\textsuperscript{1}, \;
  Chen Tang\textsuperscript{2}, \;
  Yihan Hu\textsuperscript{1}, \;
  Wei Zhan\textsuperscript{1,3} \\[4pt]
  \normalfont
  \textsuperscript{1}Applied Intuition \quad
  \textsuperscript{2}UCLA \quad
  \textsuperscript{3}UC Berkeley
}

\begin{document}
\maketitle

\begin{abstract}
End-to-end autonomous driving has achieved state-of-the-art performance on benchmarks and real-world deployments.
Its standard training recipe, however, is expensive across all stages: 
collecting and labeling millions of driving frames is costly, and closed-loop RL on images is bottlenecked by the per-step cost of photorealistic rendering plus a forward pass through a large vision backbone. 
Self-play in vectorized simulators changes the economics: millions of rollout steps per second, and a state distribution naturally rich in collisions, near-misses, and recoveries that no driving log contains.
Our approach exploits this asymmetry by decoupling \emph{learning to drive} from \emph{learning to see}. 
We \emph{pretrain} a single policy by self-play, then align its latent space with a pretrained vision backbone, through the action KL divergence and a batch-relational low-rank structural loss.
The action target comes from the self-play policy, so alignment never supervises against a logged trajectory: a paired dataset of (image, scene-state) frames suffices, with no need for the curated expert demonstrations that imitation pretraining is built on.
On photorealistic 3D Gaussian splatting closed-loop scenarios, the resulting end-to-end policy matches or exceeds prior end-to-end methods.
Project page: \url{https://terra-applied.github.io/TerraTransfer/}.
\end{abstract}



\section{Introduction}
\label{sec:intro}

Autonomous driving has received a surge of interest in the last decade, spurred by both academic research and industrial commercialization~\citep{paden2016survey,sun2020waymo,caesar2021nuplan}.
Among the many proposed solutions, end-to-end driving, which maps raw sensor inputs~(\emph{e.g.}, camera images) directly to control commands, has been widely adopted for its simplicity and effectiveness unlocked by the deep learning revolution and the availability of large-scale labeled driving data~\citep{bojarski2016end,codevilla2019cilrs,caesar2020nuscenes,sun2020waymo,caesar2021nuplan}.
It now attains state-of-the-art performance across a range of benchmarks and real-world deployments~\citep{uniad,vad,ltf,navsim,hugsim,eco}.

This capability, however, comes at a steep price.
The \defacto training recipe proceeds in several stages: 
large-scale imitation pretraining on logged human driving, followed by some form of fine-tuning, either supervised adaptation on curated demonstrations to match the deployment domain~\citep{codevilla2019cilrs,uniad,vad,ltf}, open-loop reinforcement learning~(RL) against expert trajectories to inject prior knowledge that shapes the policy's behavior in addition to the human demonstrations~\citep{karkus2025beyond}, or closed-loop RL in a sensor-rendering simulator to counter covariate shift, in which compounding errors drive the policy into states that deviate from the demonstration distribution~\citep{dosovitskiy2017carla,chen2019lbc,zhang2021roach,wu2022tcp,ltf,karkus2025beyond}.
All these stages are expensive. 
Imitation pretraining at production scale rests on expert demonstrations. The \sinequanon of the recipe, collection of multi-camera logs on the order of hundreds of thousands of hours, is bottlenecked by fleet operation, sensor calibration, and 3D label annotation~\citep{caesar2020nuscenes,sun2020waymo,caesar2021nuplan,karkus2025beyond}.
Each fine-tuning variant inherits a comparable burden: supervised fine-tuning and open-loop RL both demand carefully curated, high-quality, rare-event-heavy in-domain demonstrations, while closed-loop image RL adds a compute bottleneck, rendering each observation photorealistically and pushing it through a heavy vision backbone at every environment step~\citep{dosovitskiy2017carla,chen2019lbc,zhang2021roach,wu2022tcp,ltf,hugsim,karkus2025beyond}.

Recently, self-play has emerged as a markedly cheaper route to driving competence.
Through efficient, large-scale simulation in vectorized state spaces, it attains robust and naturalistic driving without expert demonstrations, eliminating the need for expensive human-driving data and photorealistic rendering~\citep{cusumano2025robust,chang2025spacer,seong2025grbo,guo2026correctionplanner}.
It is also well suited to the long tail: the multi-agent distribution it induces is naturally rich in collisions, near-misses, and recoveries that logged traffic seldom contains.
Together, these properties raise the prospect of learning robust, naturalistic policies with no human in the loop, at a cost orders of magnitude below the conventional workflow.
Yet self-play, for all its promise, does not yield a deployable end-to-end driving policy \perse: it consumes vectorized scene state rather than raw sensors, and so has remained largely confined to traffic simulation~\citep{cusumano2025robust,chang2025spacer,seong2025grbo,guo2026correctionplanner,konstantinidis2025toward,ahmadi2026rlftsim}.

\begin{figure}[t]
  \centering
  \begin{subfigure}[t]{0.48\linewidth}
    \centering
    \vbox to 0.63\linewidth{\vfil\includegraphics[width=\linewidth]{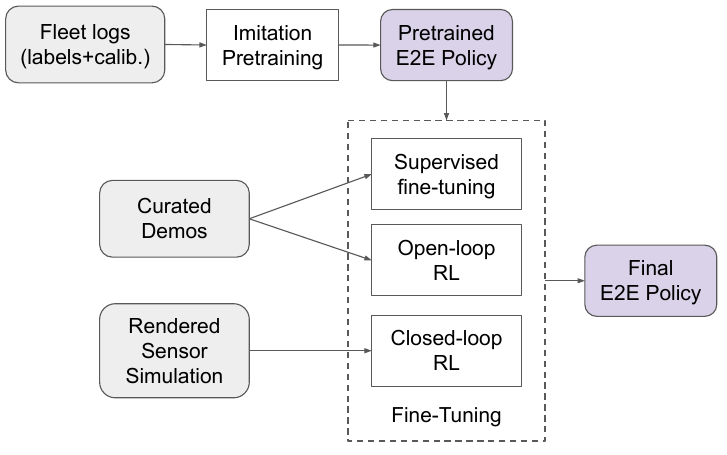}\vfil}
    \caption{\textbf{Conventional end-to-end training workflow}}
  \end{subfigure}
  \hfill
  \begin{subfigure}[t]{0.48\linewidth}
    \centering
    \vbox to 0.63\linewidth{\vfil\includegraphics[width=\linewidth]{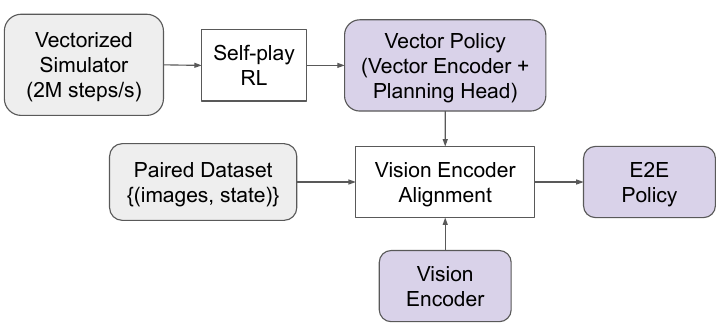}\vfil}
    \caption{\textbf{Proposed workflow}}
  \end{subfigure}
  \caption{%
    \textbf{Conventional vs.\ proposed training paradigm.}
    \textbf{(a)}~Conventional recipes begin with imitation pretraining on
    fleet-scale logs, then add supervised fine-tuning, open-loop RL on logged
    trajectories, or closed-loop image RL in sensor simulators, each path
    requiring expensive human-driving data or photorealistic rendering.
    \textbf{(b)}~Our two-phase paradigm decouples \emph{learning to drive}
    from \emph{learning to see}: Phase~1 trains a teacher policy by self-play
    via efficient simulation (2M+ steps/s);
    Phase~2 aligns a vision student to the frozen teacher on a paired
    (image, scene-state) dataset via action-distribution and structural
    feature losses, both without any expert demonstrations.}
  \label{fig:teaser}
\end{figure}

In this work, we unite the strengths of self-play and end-to-end driving in a single paradigm that learns driving policies without expert demonstrations, exploiting the cost and modality asymmetry between vector-state self-play RL and demonstration-dependent end-to-end recipes.
This is motivated by the fact that the optimal action in a specific situation depends on the scene state, not on the modality used to perceive it or on whether the action at that state was produced by a human.
For this, we propose to break the observation dependency by decoupling \emph{learning to drive} from \emph{learning to see}, where no stage of the pipeline is supervised against logged human driving.
We first train a planning head \tabularasa by multi-agent self-play in \terrazero~\citep{wu2026terrazero}, our in-house vectorized driving simulator, in the spirit of GigaFlow~\citep{cusumano2025robust}: a single parameter set controls every agent in every scene, and the head learns from the long-tail distribution of its own behavior, never imitating a human trajectory.
We then freeze this head and train an image-conditioned encoder atop a pretrained vision backbone~\citep{simeoni2025dinov3} to replace its set-valued road and partner inputs, supervised by the frozen head's per-state action distribution together with a feature-level \emph{structural} loss.

Compared to the canonical end-to-end recipe, our framework is remarkably frugal in both data and compute, yet preserves the naturalistic driving competence inherent to self-play, and, as we show, matches or exceeds prior end-to-end methods on aggregate closed-loop HD-Score over photorealistic evaluation benchmarks.
The cost of learning to drive moves upstream into self-play, but this one-time compute is efficient and reusable:
the planning head is trained in 96 hours on 16 A100 GPUs (over 2.4B kilometers of simulated driving) with no human demonstrations, and can thereafter supply alignment supervision for a range of vision frontends, alignment datasets, sensor stacks, and downstream specializations.
During alignment, the only signal the vision encoder receives is the frozen head's per-state action distribution on a paired dataset of (image, scene-state) frames, a strictly weaker requirement than collecting and labeling the expert demonstrations a behavior-cloning loss would learn from.
In our case, the encoder is aligned over 1.83M paired frames on 8 A100 GPUs in 10 hours.
Figure~\ref{fig:teaser} contrasts our paradigm with the conventional workflow.

In summary, our contributions in this work include the following:
\begin{itemize}
  \item A demonstration-free recipe for end-to-end driving: a planning head trained by self-play in a vectorized simulator, followed by image-conditioned encoders that learn to consume raw images---with no logged-trajectory supervision at any stage.
  \item A batch-relational low-rank structural loss that matches the pairwise scene similarities the teacher induces within its low-rank feature subspace, motivated by the empirical finding that the teacher's features are low-rank and redundant (\S\ref{sec:exp-rank}).
  \item Closed-loop results on photorealistic 3D Gaussian splatting scenarios~\citep{hugsim} showing that the resulting end-to-end policy matches or exceeds prior end-to-end methods---with an alignment phase that needs no expert demonstrations or logged-trajectory labels, only paired (image, scene-state) frames and the frozen teacher's action distribution.
\end{itemize}


\section{Related Work}
\label{sec:related}

\paragraph{Learning to Drive: Imitation vs.\ Self-Play.}
End-to-end driving policies are predominantly trained by imitation on logged human driving, optionally followed by fine-tuning---supervised fine-tuning on curated demonstrations, open-loop RL against logged trajectories, or closed-loop RL in a rendered simulator; UniAD~\citep{uniad} and VAD~\citep{vad} jointly supervise perception, prediction, and planning from multi-camera video, and SMART~\citep{wu2024smart} casts motion prediction as next-token generation over agent trajectories.
All such variants ultimately depend on logged human driving---directly for imitation, SFT, and open-loop RL, and indirectly for closed-loop RL through the imitation-pretrained initialization and the logged scenes used to reconstruct the simulator---so they suffer covariate shift on the rare, safety-critical states absent from the demonstration distribution and stay bounded by expert quality~\citep{karkus2025beyond}.
The closed-loop variant, typically privileged-teacher distillation in CARLA~\citep{dosovitskiy2017carla,chen2019lbc,zhang2021roach,wu2022tcp,ltf}, stays rare because each step requires photorealistic rendering and a heavy vision-backbone pass, only recently eased by 3D Gaussian splatting simulators~\citep{hugsim,hess2025splatad,gao2025rad}.
A parallel line of work sidesteps this dependence with multi-agent self-play in vectorized, object-level simulators---world state as bounding boxes, lane polylines, and partner kinematics rather than pixels---reaching high-fidelity driving without any human demonstrations: GigaFlow~\citep{cusumano2025robust} from 1.6 billion simulated kilometers, SPACeR~\citep{chang2025spacer} via KL-anchoring to a reference, \citet{seong2025grbo} via group relative policy optimization, and CorrectionPlanner~\citep{guo2026correctionplanner} via autoregressive self-correction, all underpinned by efficient simulators~\citep{suarez2024pufferlib,konstantinidis2025toward,ahmadi2026rlftsim} and echoing self-play's success in drone racing~\citep{kaufmann2023champion} and open-ended program evolution~\citep{kumar2026drq}.
These methods consume vectorized scene state, however, and none has been extended end-to-end to image observations at the scales they reach in vector form---the gap our work closes.

\paragraph{Policy Distillation and Cross-Modal Alignment.}
Aligning representations across modalities is well-studied---CLIP~\citep{radford2021clip}, ALIGN~\citep{jia2021align}, and BLIP-2~\citep{li2023blip2} align vision with language---as is policy distillation in RL, where a student reproduces teacher policies for compression or multi-task consolidation~\citep{rusu2016policy,parisotto2016actormimic,teh2017distral,schmitt2018kickstarting}.
Closest to us is privileged teacher--student training, where a teacher on privileged simulator state is distilled into a student restricted to onboard observations---across legged locomotion~\citep{lee2020learning,miki2022learning,kumar2021rma}, agile flight~\citep{loquercio2021learning}, in-hand manipulation~\citep{chen2021system}, off-road driving~\citep{wu2026tadpo}, and CARLA~\citep{chen2019lbc,zhang2021roach,wu2022tcp,ltf}---with knowledge distillation~\citep{hinton2015distilling} supplying the soft-label tooling.
Within driving, planner-centric methods supervise intermediate representations by their downstream effect on planning rather than module-level metrics: PKL~\citep{philion2020pkl} and TIP~\citep{li2023tip} score perception through a frozen planner, and LTF~\citep{ltf} distills a frozen planning head into a vision student.
These teachers, however, are scripted or privileged-state experts whose students are cloned on actions or hand-designed targets by behavior cloning or DAgger-style relabeling.
In contrast, our teacher is a self-play policy that never imitates, and the student is aligned by a relational structural loss~\citep{tung2019similarity} rather than by cloning teacher actions.


\section{Method}
\label{sec:method}

\begin{figure}[t]
\centering
\includegraphics[width=0.8\linewidth]{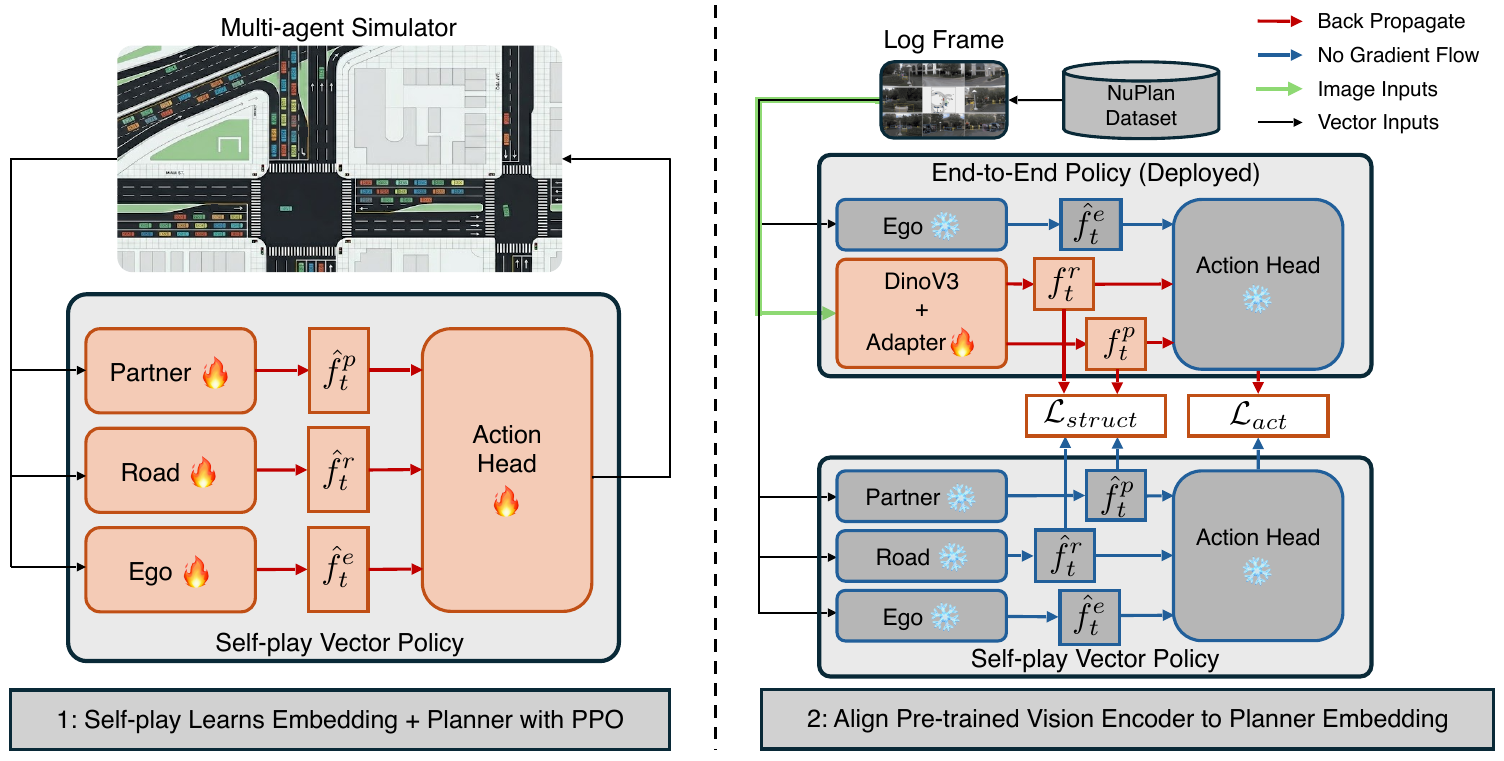}
\caption{\textbf{Two-phase training pipeline.} Flame icons mark trainable modules and snowflakes mark frozen ones; red arrows carry gradients and blue arrows do not. \textbf{Phase~1 (left):} A single self-play vector policy is trained end-to-end with PPO in a multi-agent vectorized simulator. The ego, map, and partner encoders, and the action head are jointly optimized, and the same parameter set controls every agent in the scene so the interaction distribution co-evolves with the policy. \textbf{Phase~2 (right):} The Phase-1 policy is frozen (bottom) and used purely as a source of per-frame supervision for the deployed end-to-end policy (top). For each log frame drawn from the nuPlan dataset, the student receives the camera image plus ego kinematics and a navigation signal: a frozen DINOv3 backbone extracts image features, and two linear adapters map them to the road and partner features, while the ego encoder is inherited and kept frozen. Two losses tie the student to the teacher: a \emph{structural loss} on the joint embedding and an \emph{action loss} on the output action distribution. No logged ego trajectory is ever used as a supervision target.}
\label{fig:overview}
\end{figure}

\subsection{Self-Play Pretraining}
\label{sec:method:selfplay}

\paragraph{Setup.} We pretrain a driving policy $\pi_\theta$ by self-play in \terrazero. A single set of parameters $\theta$ controls all agents in every scene; all transitions are pooled into one on-policy buffer. Each agent's ``surrounding traffic'' is therefore other instances of the current policy, so the multi-agent interaction distribution evolves with the policy itself.

\paragraph{Policy architecture.} At each step an agent observes its own ego state $o^{e}_t$ (its kinematics together with the per-episode reward-weight vector $w^{(n)}$ introduced below), a variable-sized set of road elements $\mathcal{O}^{r}_t=\{o^{r}_{t,i}\}_{i=1}^{M_r}$, and a variable-sized set of partner agents $\mathcal{O}^{p}_t=\{o^{p}_{t,j}\}_{j=1}^{M_p}$. Three encoders process these inputs in parallel:
\begin{equation}
\hat{f}^{e}_t = E_{\text{ego}}(o^{e}_t),\quad
\hat{f}^{r}_t = E_r(\mathcal{O}^{r}_t),\quad
\hat{f}^{p}_t = E_p(\mathcal{O}^{p}_t),
\label{eq:encoders}
\end{equation}
where $E_{\text{ego}}$ is an MLP and $E_r, E_p$ are DeepSets~\citep{zaheer2017deep} encoders, $E_r(\mathcal{O}^{r}_t) = \mathrm{maxpool}_i\,\phi_r(o^{r}_{t,i}) + b_r$ and likewise for $E_p$, whose learned biases $b_r, b_p$ provide a well-defined output when the corresponding set is empty. The three features are concatenated and fed through a shared MLP and an actor head to produce a categorical distribution over a discrete action set $\mathcal{A}$ (see Appendix~\ref{sec:supp-vehicle-dynamics}):
\begin{equation}
\pi_\theta(\cdot \mid o_t) = \mathrm{softmax}\!\Big(W_a\cdot \mathrm{MLP}_{\text{shared}}\big([\hat{f}^{e}_t,\,\hat{f}^{r}_t,\,\hat{f}^{p}_t]\big)\Big).
\label{eq:actor}
\end{equation}

\paragraph{Reward.} The per-step reward $R^{(n)}_t$ for agent $n$ is a weighted sum of $K$ driving-quality terms (goal-reaching, collision, comfort, lane alignment, lane centering, velocity, traffic-rule compliance, etc.; see Appendix~\ref{sec:supp-rewards}):
\begin{equation}
R^{(n)}_t \;=\; \sum\nolimits_{k=1}^{K} w^{(n)}_k \cdot r^{(k)}_t\!\big(o^{(n)}_t,\,a^{(n)}_t\big).
\label{eq:reward}
\end{equation}
Following the domain-randomization scheme of GigaFlow~\citep{cusumano2025robust}, the weights $w^{(n)}_k\sim\mathcal{U}(w_k^{\min},w_k^{\max})$ are resampled per agent per episode and exposed to the policy as part of its ego observation $o^{e}_t$, so a single $\pi_\theta$ is conditioned on, and trained to act consistently across, a continuum of driver preferences.

\paragraph{Objective.} We optimize the clipped PPO~\citep{schulman2017proximal} surrogate with a value loss and entropy bonus:
\begin{equation}
\mathcal{L}(\theta) = -\mathbb{E}_t\!\Big[\min\!\big(\rho_t\hat{A}_t,\;\mathrm{clip}(\rho_t,\,1{-}\epsilon,\,1{+}\epsilon)\hat{A}_t\big)\Big]
\;+\; c_v\,\mathcal{L}_V(\theta) \;-\; c_H\,\mathcal{H}\!\left[\pi_\theta(\cdot\mid o_t)\right],
\label{eq:ppo}
\end{equation}
where $\rho_t=\pi_\theta(a_t\mid o_t)/\pi_{\theta_{\text{old}}}(a_t\mid o_t)$ and $\hat{A}_t$ is the GAE~\citep{schulman2015high} advantage. After training, $\pi_\theta$ serves a dual role in the next phase: it is the frozen teacher that supplies per-frame supervision, and its shared MLP and actor head are inherited verbatim as the student's planning head.

\subsection{Vision Alignment}
\label{sec:method:alignment}

The student must drive the self-play policy from images, whereas the policy was trained on vector state. The obvious bridge is a \emph{cascade}: train a perception module to recover the teacher's vector inputs---bounding boxes, lane polylines, partner kinematics---from the image, then feed them through the frozen policy. We avoid it. The teacher's set encoders pool each variable-size scene into a single fixed-dimensional feature, a lossy and many-to-one map: many detection sets yield the same feature, and only that feature drives the policy. A cascade thus solves a harder problem than necessary: recovering one detection set when we can instead predict the pooled feature the policy consumes directly. This pooled feature is the alignment target we adopt. In addition, we observe that the pooled feature's informative variation is sharply low-rank and redundant (\S\ref{sec:exp-rank}); accordingly, the \emph{structural} loss below matches these features only within their low-rank subspace rather than across all coordinates. Lastly, an \emph{action} loss additionally ties the student's policy distribution to the teacher's.

\paragraph{Setup.} The pretrained teacher $\pi^T_\theta$ is frozen. We replace its two set-valued encoders --- road and partner --- with image-conditioned counterparts $E^{\text{vis}}_r, E^{\text{vis}}_p$, while inheriting the teacher's ego encoder $E_{\text{ego}}$, shared MLP, and actor head verbatim. Both vision encoders share a single DINOv3~\citep{simeoni2025dinov3} backbone with two linear adapters (Fig.~\ref{fig:overview}); this choice gave the strongest closed-loop performance in our backbone ablation (Appendix~\ref{sec:supp-backbone-ablation}). Given an image $I_t$ paired with the teacher's vector observation $o_t$, the student and teacher features are
\begin{equation}
\big(f^{r}_t,\,f^{p}_t\big) = \big(E^{\text{vis}}_r(I_t),\,E^{\text{vis}}_p(I_t)\big),\qquad
\big(\hat{f}^{r}_t,\,\hat{f}^{p}_t\big) = \big(E_r(\mathcal{O}^{r}_t),\,E_p(\mathcal{O}^{p}_t)\big),
\label{eq:align-feats}
\end{equation}
where $E_r, E_p$ are the frozen DeepSets encoders from Eq.~\ref{eq:encoders}, so the teacher features $\hat{f}^{r}_t, \hat{f}^{p}_t$ are exactly those produced during self-play, while the student's $f^{r}_t, f^{p}_t$ are their vision-derived counterparts. Passing either feature pair through the shared frozen head yields a policy distribution $\pi^S(\cdot\mid I_t)$ or $\pi^T(\cdot\mid o_t)$. We supervise the student with two objectives.

\paragraph{Structural alignment.}
Rather than forcing the student to memorize the teacher's absolute feature coordinates, we train it to reproduce the teacher's \emph{relational structure}---how the teacher arranges different driving scenes relative to one another~\citep{tung2019similarity}. Furthermore, as shown in \S\ref{sec:exp-rank}, the informative variation in the teacher's features is sharply low-rank. Matching all absolute coordinates directly would undesirably force the student to fit the structureless tail---the low-variance directions that an optimal low-rank denoiser discards as noise~\citep{gavish2014optimal}. Therefore, we align the features strictly within their principal subspace.

Formally, for a batch of size $B$ and each encoder $*\in\{r,p\}$, we first mean-center the student and teacher features into matrices $\bar{F}_*, \bar{\hat{F}}_* \in \mathbb{R}^{B\times d}$. To extract the informative subspace, we compute a per-batch basis using the top-$k_*$ right singular vectors of the stop-gradient teacher matrix:
$$
\bar{\hat{F}}_* \overset{\mathrm{SVD}}{=} U_*\Sigma_* V_*^\top,
\qquad
V_*^{(k_*)} = \big[\,v_{*,1},\dots,v_{*,k_*}\,\big]\in\mathbb{R}^{d\times k_*}.
$$
This basis $V_*^{(k_*)}$ captures the dominant axes of variation across the current scenes. We then project both the student and teacher features onto this subspace, obtaining $Z_* = \bar{F}_* V_*^{(k_*)}$ and $\hat{Z}_* = \bar{\hat{F}}_* V_*^{(k_*)}\in\mathbb{R}^{B\times k_*}$.

Next, we capture the relational geometry by computing the cosine-similarity matrix among the $B$ scenes for both sides:
$$
S_* = \tilde{Z}_*\,\tilde{Z}_*^\top, \quad
\hat{S}_* = \tilde{\hat{Z}}_*\,\tilde{\hat{Z}}_*^\top \in \mathbb{R}^{B\times B},
\qquad
\tilde{Z}_{*,i} = \frac{Z_{*,i}}{\lVert Z_{*,i}\rVert}.
$$
Here, $\hat{S}_{*,ij}$ encodes the angle between scenes $i$ and $j$ in the teacher's subspace---effectively representing which scenes the teacher treats as alike. The student learns to replicate this geometry by minimizing the Frobenius norm of their difference:
$$
\mathcal{L}^{(*)}_{\text{struct}} \;=\; \frac{1}{B^2}\big\lVert S_* - \hat{S}_* \big\rVert_F^2,
\qquad
\mathcal{L}_{\text{struct}} = \mathcal{L}^{(r)}_{\text{struct}} + \mathcal{L}^{(p)}_{\text{struct}}.
$$
Because the basis $V_*^{(k_*)}$ and target matrix $\hat{S}_*$ are recomputed per batch under a stop-gradient, the student focuses purely on matching pairwise scene relationships within the active subspace, leaving both the absolute coordinate frame and the orthogonal noise complement free. We use $k_p=13$ and $k_r=9$ (the $80\%$ cumulative-energy cutoffs of the partner and road teacher spectra, \S\ref{sec:exp-rank}), and ablate this subspace rank alongside a full-coordinate baseline in Appendix~\ref{sec:supp-loss-ablation}.

\paragraph{Action alignment.} Beyond the features, we match the student's action distribution to the teacher's by forward KL divergence,
\begin{equation}
\mathcal{L}_{\text{act}} \;=\; \mathrm{KL}\!\big(\pi^T(\cdot\mid o_t)\;\|\;\pi^S(\cdot\mid I_t)\big),
\label{eq:loss-act}
\end{equation}
which provides a global, policy-level consistency signal: whatever the student does to its road/partner features, the resulting actor logits must end up close to the teacher's.

\paragraph{Total loss.} The alignment phase optimizes: $\mathcal{L}_{\text{align}} \;=\; \mathcal{L}_{\text{act}} \;+\; \lambda\,\mathcal{L}_{\text{struct}}$. We set $\lambda=0.5$, to which performance is largely insensitive (Appendix~\ref{sec:supp-loss-ablation}).


\section{Experiments}
\label{sec:experiments}

We evaluate whether a vision policy aligned to a self-play teacher, with no
supervision from any logged trajectory, can drive competitively in closed loop.
Our headline comparison (\S\ref{sec:exp-headline}) shows that the aligned vision
student matches or outperforms imitation-trained end-to-end pipelines and comes
close to its own teacher, even though it is aligned on \nuplan and evaluated on
\nuscenes-derived scenarios. We then quantify how little paired data this takes
(\S\ref{sec:exp-data-efficiency}): with no trajectory labels and roughly $40\%$
of the \nuplan frames that the state-of-the-art baseline fine-tunes
on, the student already surpasses it. Finally, we trace the method back to a property
of the teacher (\S\ref{sec:exp-rank}): its features are low-rank, which is what
makes matching only a relational low-rank subspace an effective alignment
target. Remaining design choices and detail-level ablations are deferred to the
appendix: the loss ablation (Appendix~\ref{sec:supp-loss-ablation}), the
vision backbone (Appendix~\ref{sec:supp-backbone-ablation}), and the sensitivity of alignment to
the rollout policy used to collect paired frames
(Appendix~\ref{sec:supp-demo-quality}).

\subsection{Setup}
\label{sec:exp-setup}
We evaluate on \hugsim~\citep{hugsim}, a photorealistic 3D Gaussian splatting
closed-loop benchmark built from \nuscenes-derived scenarios and split into
Easy, Medium, Hard, and Extreme tiers. We report the benchmark's aggregate
\hdscore; metric definitions are provided in Appendix~\ref{sec:supp-cl-hdscore}.

Because our policy emits one immediate action at a time, the native open-loop
time-to-collision (\ttc) and comfort (\comscore) definitions, which assume an
explicitly planned future trajectory, are not directly applicable. We therefore
use closed-loop versions of these two metrics computed from the realized
rollout, while leaving the rest of the \hdscore computation unchanged.
Appendix~\ref{sec:supp-cl-hdscore} details this conversion;
Appendix~\ref{sec:supp-self-play-policy-baseline} provides a more detailed
evaluation of the self-play teacher policy.

\subsection{Closed-Loop Comparison with End-to-End Baselines}
\label{sec:exp-headline}

\begin{table}[ht]
\caption{\textbf{Our vision policy approaches the self-play teacher and outperforms
published end-to-end baselines on \hugsim.}
Closed-loop \hdscore (\S\ref{sec:exp-setup}) across the four
\hugsim~\citep{hugsim} difficulty tiers on the 88 \nuscenes-derived scenarios. Higher
is better; \textbf{bold} marks the best per column among comparison methods. The
\textit{italicized} self-play baseline is the teacher we distill from---unlike
every other row, it drives from privileged vectorized scene state rather than
camera images---included as a reference and excluded from best-per-column counts.}
\label{tab:hugsim-headline}
\centering
\small
\renewcommand{\arraystretch}{1.15}
\setlength{\tabcolsep}{6pt}
\begin{tabular}{@{}lccccc@{}}
\toprule
Method & Easy & Medium & Hard & Extreme & All \\
\midrule
\uniad~\citep{uniad}                 & 0.367 & 0.198 & 0.249 & 0.109 & 0.224 \\
\vad~\citep{vad}                     & 0.400 & 0.228 & 0.242 & 0.095 & 0.239 \\
\ltf~\citep{ltf}                     & 0.634 & 0.391 & 0.289 & 0.098 & 0.360 \\
\eco Smoothing-only~\citep{eco}      & 0.764 & 0.416 & 0.405 & \textbf{0.255} & 0.452 \\
\eco Smoothing + Re-time~\citep{eco} & 0.720 & 0.388 & 0.342 & 0.236 & 0.415 \\
\midrule
\textit{Self-play baseline (ref.)}   & \textit{0.780} & \textit{0.497} & \textit{0.639} & \textit{0.185} & \textit{0.520} \\
\textbf{Ours}                        & \textbf{0.769} & \textbf{0.501} & \textbf{0.560} & 0.150 & \textbf{0.490} \\
\bottomrule
\end{tabular}
\end{table}

Table~\ref{tab:hugsim-headline} compares closed-loop \hdscore on the \hugsim
\nuscenes scenarios. Our aligned vision policy is the strongest method on
aggregate: it reaches $0.490$, outperforming the best imitation-trained
end-to-end baseline, \ltf, by $0.130$, and the strongest published \eco variant
by $0.038$. It is also close to the self-play teacher ($0.520$), indicating that
the vision alignment transfers most of the teacher's driving competence without
logged-trajectory supervision.

The main exception is Extreme, where \eco leads by $0.105$. These scenarios are
strongly out of distribution: surrounding vehicles may actively collide with the
ego, and in some cases pass through other vehicles before doing so. Our policy
responds conservatively, which preserves safety behavior but often sacrifices
route completion; the resulting low \rc pulls down the aggregate \hdscore on
this split.

\subsection{Alignment Data Efficiency}
\label{sec:exp-data-efficiency}

The paired dataset is needed only to teach the vision frontend how to reproduce
the frozen teacher's behavior. The headline run uses $\approx\!1.83$M paired
(image, scene-state) frames from \nuplan~\citep{caesar2021nuplan}
($7{,}129$ shards of $256$ frames; $\sim$51 hours at $10$\,Hz). At every frame,
the target is the teacher's action distribution on the reconstructed scene; the
logged ego trajectory is never used. These frames come from \nuplan, whereas
\hugsim is built from \nuscenes scenarios (\S\ref{sec:exp-setup})---distinct
datasets with different sensor rigs and locations---so the closed-loop scores
here additionally reflect cross-dataset generalization rather than
in-distribution replay.

\begin{figure}[ht]
\hspace{-0.05\linewidth}
\centering
\captionsetup{font=small,skip=4pt}
\begin{minipage}[t]{0.6\linewidth}
\centering
\includegraphics[width=0.96\linewidth]{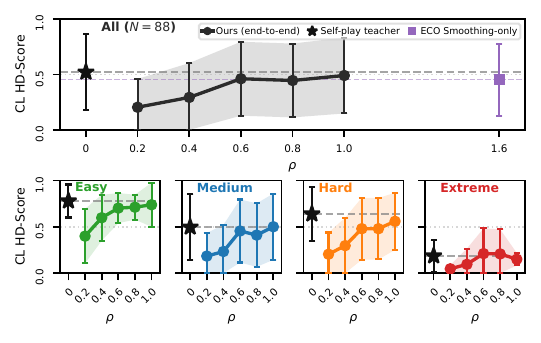}
\caption{\textbf{Alignment data efficiency.} Closed-loop \hdscore \vs relative \nuplan training data $\rho$ (our full alignment set $=1.83$M frames $\Rightarrow \rho\!=\!1$), for the All set (top) and each \hugsim tier (bottom) on \nuscenes; bands are $\pm 1$ per-scene std. The self-play teacher (\S\ref{sec:exp-headline}) uses no \nuplan data and sits at $\rho\!=\!0$ (horizontal dashed line) in every panel; \eco Smoothing-only is placed at $\rho\!\approx\!1.6$ ($\sim$2.8M \nuplan frames \vs our 1.83M), All set only, for reference.}
\label{fig:data-efficiency}
\end{minipage}\hfill
\begin{minipage}[t]{0.42\linewidth}
\centering
\includegraphics[width=0.9\linewidth]{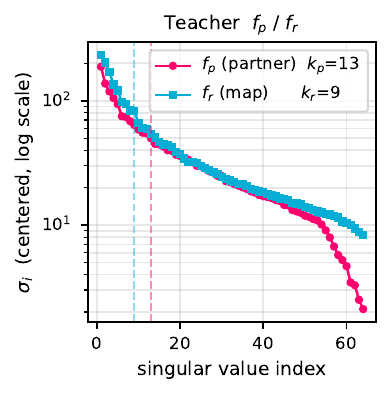}
\caption{\textbf{Singular value spectra of the teacher features.} Centered \svd of teacher $f_p$ (partner) and $f_r$ (map) over the alignment manifest. Dashed lines mark the $80\%$ cumulative-energy index ($k_p\!=\!13$ partner, $k_r\!=\!9$ map), the truncations used in \S\ref{sec:method:alignment}.}
\label{fig:svd-latent-rank}
\end{minipage}
\end{figure}

Figure~\ref{fig:data-efficiency} sweeps the paired-data fraction
$\rho\in\{0.2,0.4,0.6,0.8,1.0\}$ of our $1.83$M frames while holding the
alignment recipe fixed. For reference it also places \eco, the strongest
published baseline, at $\rho\approx1.6$: on top of large-scale imitation
pretraining, \eco fine-tunes on the full \nuplan training split
($\sim$2.8M frames), well beyond our budget. Our student needs far less
in-domain data and no trajectory labels: it already surpasses \eco at
$\rho=0.6$ ($0.461$ \vs $0.452$, roughly $40\%$ of \eco's \nuplan frames), stays
around that level at $\rho=0.8$ ($0.445$), and reaches $0.490$ at full data,
surpassing \eco by $0.038$.

\subsection{Effective Rank of Latent Features}
\label{sec:exp-rank}

Figure~\ref{fig:svd-latent-rank} explains why the structural loss aligns a
low-rank subspace rather than all feature coordinates. The teacher's map and
partner features live in a much smaller effective space than their
$D=64$-dimensional parameterization: the map spectrum captures $80\%$ of its
energy in the first $9$ directions, and the partner spectrum in the first $13$.
The entropy-based effective ranks~\citep{roy2007effective} are similarly low,
about $15$ for map and $18$ for partner.

This makes the rank truncation in \S\ref{sec:method:alignment} an empirical
choice rather than a free hyperparameter. We set $k_r=9$ and $k_p=13$ to retain
the structured part of the teacher representation while avoiding the long tail
of small singular directions, where matching absolute coordinates would mostly
ask the vision student to fit weakly structured variation.
Appendix~\ref{sec:supp-loss-ablation} ablates this low-rank subspace loss
against a full-rank variant to confirm the truncation on 80\% energy helps closed-loop
performance.


\vspace{-6pt}
\section{Limitations}
\label{sec:limitations}
\vspace{-4pt}
Our current approach faces three main limitations stemming from the simulated training environment. First, training solely on bounding boxes means the policy misses fine-grained visual cues, such as brake lights or hand signals, which we plan to resolve through supervised fine-tuning on real-world data. Second, because the simulator currently uses only vehicle kinematics models, the policy has limited exposure to mixed traffic involving pedestrians and cyclists. Finally, incomplete support for signalized intersections means the system has yet to learn complex right-of-way rules. We expect to mitigate these simulation constraints in future work by expanding the environment's supported entities and traffic logic.


\vspace{-6pt}
\section{Conclusion}
\label{sec:conclusion}
\vspace{-4pt}

We presented a demonstration-free recipe for end-to-end driving: instead of imitating logged human trajectories, we learn to drive once by self-play in \terrazero, our vectorized simulator, and then transfer that competence to a vision policy in a single alignment stage. Phase~1 trains a planning head by multi-agent self-play on vector state. Phase~2 freezes it and aligns a vision frontend to it on paired (image, scene-state) frames, supervised by an action-distribution loss and a batch-relational low-rank structural loss, never against a logged trajectory. The resulting raw-image policy comes within $0.03$ aggregate HD-Score of its self-play teacher and outperforms imitation-trained end-to-end pipelines, edging out the strongest published baseline on aggregate, on photorealistic closed-loop scenarios. The recipe moves the cost of learning to drive upstream into self-play, which is demonstration-free, scalable, one-time, and reusable. In essence, this work highlights simulation-driven self-play as a highly effective, data-efficient path forward for end-to-end autonomous driving.

\clearpage

\bibliography{ref}


\appendix

\clearpage
\section*{Supplementary Contents}
\vspace{0.8em}
\noindent\hyperref[sec:supp-self-play-details]{A\quad Self-Play Policy Details}\dotfill \pageref{sec:supp-self-play-details}\par
\noindent\hyperref[sec:supp-cl-hdscore]{B\quad Closed-Loop HD-Score}\dotfill \pageref{sec:supp-cl-hdscore}\par
\noindent\hyperref[sec:supp-self-play-policy-baseline]{C\quad Self-Play Policy Baseline Performance}\dotfill \pageref{sec:supp-self-play-policy-baseline}\par
\noindent\hyperref[sec:supp-loss-ablation]{D\quad Loss Ablation}\dotfill \pageref{sec:supp-loss-ablation}\par
\noindent\hyperref[sec:supp-backbone-ablation]{E\quad Vision Backbone Ablation}\dotfill \pageref{sec:supp-backbone-ablation}\par
\noindent\hyperref[sec:supp-demo-quality]{F\quad Alignment Decoupled from Demonstration Quality}\dotfill \pageref{sec:supp-demo-quality}\par
\noindent\hyperref[sec:supp-qualitative]{G\quad Qualitative Results}\dotfill \pageref{sec:supp-qualitative}\par

\clearpage

\section{Self-Play Policy Details}
\label{sec:supp-self-play-details}

\subsection{Training Details}
\label{sec:supp-training-details}

\paragraph{Environment setup.}
Self-play training runs in \terrazero, our in-house vectorized driving
simulator, and uses \nuplan HD maps as map geometry only. We compile
\nuplan map geometry into simulator binaries and use lane topology, lane
boundaries, and drivable-area polygons as static scene context; we do
\emph{not} use \nuplan logged trajectories, logged agent initial states,
logged goals, human actions, or replayed traffic as training supervision.
Controlled vehicles are spawned procedurally from lane centerlines by
rejection sampling, so the initial scene is collision-free and drivable. The
learned policy controls vehicles only; the
simulator also procedurally places vulnerable road users and static obstacles
as scene context, but these entities are not trained policies. Goals are also
procedural. For each spawned vehicle, the simulator samples a target distance
between $20$ and $100$\,m and walks forward along the HD-map lane graph,
selecting reachable lane successors and filtering candidates by connectivity
and heading consistency. When a vehicle reaches its current goal, a new
reachable goal is sampled from the same topology; no destination is taken from
a log.

\paragraph{RL algorithm.}
The teacher is trained as a decentralized policy with PPO. We use
V-trace-corrected GAE ($\gamma=0.99$, $\lambda=0.95$), clipped PPO with clip
coefficient $0.2$, entropy coefficient $0.01$, value coefficient $0.5$, Adam
with an annealed $5{\times}10^{-4}$ learning rate, and \texttt{bfloat16} mixed
precision.
The reference teacher run was trained on $16$ A100 GPUs, with approximately $3$M cumulative agent steps/s throughput.

\subsection{Observations}
\label{sec:supp-observations}

At each simulator step ($\Delta t=0.1$\,s), a controlled vehicle observes
three groups of inputs matching the policy factorization in
Eq.~\ref{eq:encoders}: an ego vector, a variable-size set of road elements,
and a variable-size set of nearby partners.
The ego vector contains the vehicle kinematic state
$(x,y,\theta,v,a_{\text{long}},a_{\text{lat}},\phi)$ in the ego frame,
the current navigation/goal signal, the sampled vehicle geometry and
dynamics coefficients, and the per-episode reward-weight vector.
Road observations encode nearby lane and boundary polylines relative to the
ego frame. Partner observations encode neighboring scene entities by relative
pose, velocity, heading, size, type, and validity masks. The road and partner
sets are pooled by the DeepSets encoders described in
\S\ref{sec:method:selfplay}.

\subsection{Reward Components}
\label{sec:supp-rewards}

The vehicle reward in Eq.~\ref{eq:reward} uses most of the reward components
and settings from the self-play setup of \citet{cusumano2025robust}, including
the per-agent, per-episode reward-weight randomization exposed through the ego
observation. Our only reward modification is the addition of a lane-change
reward term that penalizes excessive lane changes during self-play.
Table~\ref{tab:supp-vehicle-reward-coefficients} lists the exact coefficient
settings used in our training run.

\begin{table}[htbp]
\centering
\caption{Vehicle reward coefficients used during self-play. Ranges denote
independent per-vehicle, per-episode uniform sampling.}
\label{tab:supp-vehicle-reward-coefficients}
\footnotesize
\setlength{\tabcolsep}{3pt}
\begin{tabular}{@{}p{0.22\linewidth}p{0.39\linewidth}p{0.31\linewidth}@{}}
\toprule
Reward parameter & Coefficient setting & Effect \\
\midrule
Collision & $\alpha_{\text{collision}}\sim\mathcal{U}(0.0,3.0)$ & Collision avoidance \\
Collision speed & $\alpha_{\text{coll-speed}}=0.1$ & Impact severity \\
Boundary & $\alpha_{\text{boundary}}\sim\mathcal{U}(0.0,3.0)$ & Drivable-area compliance \\
Comfort & $\alpha_{\text{comfort}}\sim\mathcal{U}(0.0,0.1)$ & Acceleration/jerk comfort \\
Lane alignment & $\alpha_{l\text{-align}}\sim\mathcal{U}(2.5{\times}10^{-4},2.5{\times}10^{-2})$ & Lane heading \\
Velocity alignment & $\alpha_{v\text{-align}}\sim\mathcal{U}(0.0,1.0)$ & Forward progress \\
Lane centering & $\alpha_{l\text{-center}}\sim\mathcal{U}(2.5{\times}10^{-4},7.5{\times}10^{-3})$ & Lane position \\
Reverse & $\alpha_{\text{reverse}}\sim\mathcal{U}(2.5{\times}10^{-4},7.5{\times}10^{-3})$ & Reverse motion \\
Lane change & $\alpha_{\text{lane-change}}\sim\mathcal{U}(0.0,0.1)$ & Lane-change regularization \\
Velocity & $\alpha_{\text{velocity}}=2.5{\times}10^{-3}$ & Speed shaping \\
Timestep & $\alpha_{\text{timestep}}=2.5{\times}10^{-5}$ & Per-step cost \\
Goal tolerance & $\delta_{\text{goal}}\sim\mathcal{U}(2.0,12.0)$\,m & Completion radius \\
Goal & $\alpha_{\text{goal}}=1.0$ & Goal completion \\
\bottomrule
\end{tabular}
\end{table}

\subsection{Vehicle Dynamics}
\label{sec:supp-vehicle-dynamics}

We use the jerk-actuated bicycle dynamics and the discrete $12$-action
longitudinal/lateral jerk grid from the vehicle branch of
\citet{cusumano2025robust}. As in that setup, each controlled vehicle
maintains pose, speed, longitudinal and lateral acceleration, and steering
state; selected jerk actions are integrated into accelerations, speed,
steering, and finally pose using the simulator time step.

Our dynamics setup differs from \citet{cusumano2025robust} in three ways.
First, because the learned policy controls vehicles only, dynamics
randomization is restricted to passenger-car-sized bodies rather than the
broader road-user size range used in that work:
\begin{equation}
\textit{length}\sim\mathcal{U}(3.5,5.5)\,\mathrm{m},\qquad
\textit{width}\sim\mathcal{U}(1.5,2.5)\,\mathrm{m},\qquad
\textit{height}\sim\mathcal{U}(1.2,2.0)\,\mathrm{m}.
\label{eq:supp-size-randomization}
\end{equation}
Second, at every episode reset each vehicle samples the scalar dynamics
coefficients
\begin{equation}
C_{\text{throttle}},C_{\text{steer}},C_{\text{acc}},C_{\text{vel}}
\sim\mathcal{U}(0.5,1.5),
\label{eq:supp-dyn-randomization}
\end{equation}
rather than the mixed-uniform coefficient distributions in
\citet{cusumano2025robust}. These coefficients and the sampled vehicle
dimensions are included in the ego observation so the policy can condition
on the vehicle it is controlling. Third, we run self-play with simulator
step $\Delta t=0.1$\,s, which improves closed-loop reactivity and matches the
control rate used by the student vision policy.

\section{Closed-Loop HD-Score}
\label{sec:supp-cl-hdscore}

The native HUGSim~\citep{hugsim} HD-Score is not well designed for closed-loop evaluation: it scores the policy's \emph{planned} future trajectory rather than the path the ego actually drives.\footnote{Throughout this appendix, the \emph{native} HD-Score denotes the public HUGSim scorer \texttt{sim/utils/score\_calculator.py} (repository \texttt{hyzhou404/HUGSIM}, commit \texttt{62c690d}, 2025-11-08; \url{https://github.com/hyzhou404/HUGSIM}), whose \texttt{calculate()} routine is the scoring entry point we benchmark against; all statements about native behavior below are taken from it.}
This is a legacy of the open-loop NAVSIM-style protocol~\citep{navsim} it was written for, where each keyframe carries a multi-second planned trajectory and the metric never has to commit to what is executed.
But what we want to measure is closed-loop driving quality, which lives in the \emph{realized} rollout trajectory, not in a plan the policy may never follow---and our single-step policy emits no multi-second plan to begin with.
We therefore introduce a closed-loop variant of the HD-Score (used in \S\ref{sec:exp-setup}) that scores the realized rollout directly: we keep the aggregation formula and the no-collision (NC) and drivable-area-compliance (DAC) subscores unchanged, redefine the time-to-collision (TTC) and comfort (COM) terms to operate on the realized trajectory, and reground route completion ($R_c$) in the scene geometry.

\subsection{Closed-loop redefinitions}
\label{sec:supp-cl-hdscore-defs}

Let $\mathcal{T}=\{t_1,\dots,t_M\}$ be the set of evaluation keyframes (a fixed-stride decimation of the rollout, default $0.5\,\mathrm{s}$), let $(x^e_t,y^e_t,\theta^e_t,v^e_t)$ be the realized pose and instantaneous velocity of the ego at keyframe $t$, and let $\mathcal{P}_t$ index the set of valid vehicle-like partners visible at keyframe $t$.
We continue to compute the aggregate as
\begin{equation}
\mathrm{HD\text{-}Score}\;=\;\frac{1}{|\mathcal{T}|}\sum_{t\in\mathcal{T}}\mathrm{NC}_t\cdot\mathrm{DAC}_t\cdot\frac{5\,\mathrm{TTC}_t+2\,\mathrm{COM}_t}{7}\;\cdot\;R_c,
\label{eq:supp-cl-hdscore-agg}
\end{equation}
with NC, DAC defined identically to native HUGSim and TTC, COM, $R_c$ redefined below.

\paragraph{Closed-loop TTC.}
We discard the planned trajectory and measure imminent risk directly on the realized rollout, addressing all three issues above.
At keyframe $t$ we propagate the ego \emph{and} every valid partner $j\in\mathcal{P}_t$ forward by their observed instantaneous velocities along a short horizon $H_{\text{TTC}}$ (default $1.0\,\mathrm{s}$, the conventional imminent window and the native velocity-shift upper bound) at a fine stride $\Delta_{\text{TTC}}$ (default $0.2\,\mathrm{s}$):
\begin{equation}
(x^{(a)}_{t,\delta},\,y^{(a)}_{t,\delta}) = (x^{(a)}_t,\,y^{(a)}_t) + \delta\,(v^{(a)}_{x,t},\,v^{(a)}_{y,t}),\quad a\in\{\mathrm{ego}\}\cup\mathcal{P}_t,\ \ \delta\in\{0,\Delta_{\text{TTC}},\ldots,H_{\text{TTC}}\}.
\label{eq:supp-cl-ttc-prop}
\end{equation}
At each $\delta$ we test the propagated ego oriented bounding box (OBB) against both each propagated partner box and the static 3D Gaussian-splatting (3DGS) point cloud $\mathcal{B}$ reconstructed for the scene background, the latter flagged when the ego box encloses more than $N_{\text{bg}}$ such points (default $100$, matching the native background-collision test), and record the earliest collision time
\begin{equation}
\tau^\star_t \;=\; \min\!\left\{\delta\;:\;\Big(\exists\,j\in\mathcal{P}_t,\;\mathrm{OBB}_t^{\text{ego}}(\delta)\cap\mathrm{OBB}_t^{(j)}(\delta)\neq\varnothing\Big)\;\vee\;\Big(\big|\mathcal{B}\cap\mathrm{OBB}_t^{\text{ego}}(\delta)\big|> N_{\text{bg}}\Big)\right\},
\label{eq:supp-cl-ttc-tau}
\end{equation}
with $\tau^\star_t = +\infty$ if no collision occurs over the horizon, and report the fractional score
\begin{equation}
\mathrm{TTC}_t \;=\; \min\!\Big(1,\,\tau^\star_t\,/\,H_{\text{TTC}}\Big)\in[0,1].
\label{eq:supp-cl-ttc}
\end{equation}
These three changes---realized poses, a fractional imminent-window sweep, and propagated (rather than frozen) partners---resolve issues~(i)--(iii) respectively.
A $\delta=0$ collision (ego already in contact at the keyframe) returns $\mathrm{TTC}_t = 0$ directly.

\paragraph{Closed-loop COM.}
We retain the native comfort bounds (the five active constraints above) but apply them to the \emph{realized} ego trajectory over a fixed-duration forward window of $W_{\text{COM}}$ seconds (default $1.0\,\mathrm{s}$) starting at each keyframe.
Concretely, at keyframe $t$ we take the forward window $\mathcal{W}_t=\{t,\,t+\Delta_{\text{sim}},\,\ldots,\,t+(n-1)\Delta_{\text{sim}}\}$ of the next $n=\max(\mathrm{round}(W_{\text{COM}}/\Delta_{\text{sim}}),\,4)$ realized ego poses ($n=10$ at the defaults $W_{\text{COM}}=1.0\,\mathrm{s}$, $\Delta_{\text{sim}}=0.1\,\mathrm{s}$, with $\Delta_{\text{sim}}$ the simulator step), compute first- and second-order finite differences to obtain longitudinal and yaw kinematics, and set $\mathrm{COM}_t = 1$ iff all of these bounds hold at every sample in $\mathcal{W}_t$ (defaulting to $1$ if fewer than four samples remain, as in native), else $\mathrm{COM}_t = 0$:
\begin{equation}
\mathrm{COM}_t \;=\;\bigwedge_{s\in\mathcal{W}_t}\Big[\,a^\parallel_s\in[-4.05, 2.40]\,\wedge\,|j^\parallel_s|\le 8.37\,\wedge\,|\alpha_s|\le 1.93\,\wedge\,|\omega_s|\le 0.95\,\Big],
\label{eq:supp-cl-com}
\end{equation}
where $a^\parallel_s$ is longitudinal acceleration ($\mathrm{m/s}^2$), $j^\parallel_s$ longitudinal jerk ($\mathrm{m/s}^3$), $\alpha_s$ yaw acceleration ($\mathrm{rad/s}^2$), and $\omega_s$ yaw rate ($\mathrm{rad/s}$).
Parameterizing the window in seconds rather than step count keeps the metric invariant to the rollout $\Delta_{\text{sim}}$.

\paragraph{Arc-length route completion.}
Each HUGSim scene supplies a densified camera-pose polyline $\mathcal{R}=(\mathbf{r}_0,\dots,\mathbf{r}_L)\subset\mathbb{R}^2$ that traces the reference route through the scene, with total arc length $S=\sum_{i=1}^{L}\|\mathbf{r}_i-\mathbf{r}_{i-1}\|$.
At each keyframe $t$ we snap the realized ego position $(x^e_t,y^e_t)$ to its nearest polyline vertex $i^\star_t=\arg\min_{i}\,\|(x^e_t,y^e_t)-\mathbf{r}_i\|$ and read off the cumulative arc length $s_t=\sum_{i=1}^{i^\star_t}\|\mathbf{r}_i-\mathbf{r}_{i-1}\|$.
The closed-loop $R_c$ is then
\begin{equation}
R_c \;=\; \min\!\Big(1,\;\max_{t\in\mathcal{T}}\,s_t\,/\,S\Big).
\label{eq:supp-cl-rc}
\end{equation}
Arc-length normalization makes $s_t/S$ a genuine distance-along-route fraction insensitive to non-uniform pose spacing, and dropping the discount makes $R_c=1$ mean the route was actually completed (fixing the two problems above).

\subsection{Native vs.\ closed-loop summary}
\label{sec:supp-cl-hdscore-table}

\begin{table}[htbp]
    \centering
    \caption{Native HD-Score vs.\ closed-loop HD-Score per subscore. NC, DAC, and the aggregation formula are unchanged; TTC, COM, and $R_c$ are redefined to operate on the realized rollout.}
    \label{tab:supp-cl-hdscore-comparison}
    \scriptsize
    \renewcommand{\arraystretch}{1.25}
    \setlength{\tabcolsep}{4pt}
    \begin{tabular}{@{}p{0.10\linewidth}p{0.40\linewidth}p{0.40\linewidth}@{}}
    \toprule
    Sub & Native HUGSim HD-Score & Closed-loop variant (this work) \\
    \midrule
    NC  & OBB ego--partner collision per keyframe; vehicle-only partner filter; output $\in\{0,1\}$. & Same. \\
    DAC & Four-corner-inside check against per-scene drivable-area polylines; output $\in\{0, 0.5, 1\}$. & Same. \\
    TTC & Entire planned trajectory ($7$ poses, $\le\!3.5\,\mathrm{s}$) rigidly shifted by ego velocity at two horizons $\Delta\in\{0.5,1.0\}\,\mathrm{s}$; collision of every shifted ego pose vs.\ \emph{frozen} partners $+$ scene geometry ($14$ probes); binary $\{0,1\}$. & Realized ego \emph{and} all valid partners velocity-propagated on a fine grid (stride $\Delta_{\text{TTC}}=0.2\,\mathrm{s}$, horizon $H_{\text{TTC}}=1.0\,\mathrm{s}$); earliest ego collision vs.\ propagated partners $+$ 3DGS background ($>\!N_{\text{bg}}$ enclosed points) gives fractional $\min(1,\,\tau^\star_t/H_{\text{TTC}})\in[0,1]$ (Eq.~\ref{eq:supp-cl-ttc}); $\delta{=}0$ collision $\Rightarrow$ TTC$_t = 0$. \\
    COM & Kinematic comfort bounds (lon.\ accel., lon.\ jerk, yaw accel., yaw rate; a lateral bound is coded but inert) on the keyframe's \emph{planned} trajectory poses; binary $\{0,1\}$. & Same active bounds on the \emph{realized} ego trajectory over a seconds-parameterized window of duration $W_{\text{COM}}=1.0\,\mathrm{s}$ (Eq.~\ref{eq:supp-cl-com}). \\
    $R_c$ & $\min(1, \max_t r_t)$ with $r_t=(k_t{+}1)/(0.9N)$: a camera-pose-\emph{index} fraction ($k_t$ = nearest pose index, $N$ = pose count), inflated by an unexplained $0.9$ discount; not physical distance. & Arc-length fraction $\min(1, \max_t s_t/S)$ along the scene's densified camera-pose polyline ($S$ = full route length, no discount): physical distance, robust to non-uniform pose spacing (Eq.~\ref{eq:supp-cl-rc}). \\
    Agg.\ & $\mathrm{mean}_t[\mathrm{NC}_t\cdot\mathrm{DAC}_t\cdot(5\,\mathrm{TTC}_t+2\,\mathrm{COM}_t)/7]\cdot R_c$. & Same (Eq.~\ref{eq:supp-cl-hdscore-agg}). \\
    \bottomrule
    \end{tabular}
\end{table}

Because the redefined TTC and COM read different motion from native (realized rollout vs.\ planned trajectory) and, for TTC, a different scale (continuous vs.\ binary), neither is directly comparable to its native counterpart per keyframe, even on the same rollout. NC and DAC are comparable. The detailed comparison is summarized in Table~\ref{tab:supp-cl-hdscore-comparison}.

\section{Self-Play Policy Baseline Performance}
\label{sec:supp-self-play-policy-baseline}

This appendix examines the self-play teacher in more detail---the privileged vector-state policy our vision student is distilled from, and the effective upper bound it is compared against in Table~\ref{tab:hugsim-headline}. We characterize it along two axes: its closed-loop driving quality on the HUGSim benchmark, broken down by difficulty tier (Table~\ref{tab:supp-hugsim-ours}), and the realism of its own rollouts in exported nuScenes traffic (Table~\ref{tab:supp-wosac-baseline}).

\begin{table}[htbp]
    \centering
    \caption{Self-play policy baseline. Per-difficulty breakdown of the closed-loop HD-Score (\S\ref{sec:supp-cl-hdscore}) for the self-play teacher on the HUGSim 88-scenario split; this is the teacher row of Table~\ref{tab:hugsim-headline}. The teacher drives each scenario from the privileged \emph{vectorized} scene state---object boxes, lane geometry, and partner kinematics, the modality it was trained on---not the rendered camera images our vision student consumes; background traffic is replayed. \emph{HD-Score} is the headline metric; the \emph{All} row aggregates over the full split.}
    \label{tab:supp-hugsim-ours}
    \scriptsize
    \renewcommand{\arraystretch}{1.18}
    \setlength{\tabcolsep}{3pt}
    \begin{tabular}{@{}lrrrrrrrr@{}}
    \toprule
    Set & N & NC & DAC & TTC & COM & HD-PDMS & $R_c$ & HD-Score \\
    \midrule
    Easy    & 18 & 0.9883 & 1.0000 & 0.9548 & 0.8926 & 0.9304 & 0.8416 & 0.7796 \\
    Medium  & 34 & 0.9690 & 1.0000 & 0.9297 & 0.8186 & 0.8874 & 0.5451 & 0.4971 \\
    Hard    & 18 & 0.9676 & 1.0000 & 0.9359 & 0.8402 & 0.8905 & 0.7124 & 0.6388 \\
    Extreme & 18 & 0.9597 & 1.0000 & 0.8315 & 0.5731 & 0.7283 & 0.2406 & 0.1851 \\
    \midrule
    All     & 88 & 0.9708 & 1.0000 & 0.9160 & 0.7879 & 0.8643 & 0.5777 & 0.5200 \\
    \bottomrule
    \end{tabular}
\end{table}

\paragraph{What limits the teacher.}
Safety and on-road compliance hold up across all four tiers---NC stays above $0.95$, DAC is saturated at $1.0$, and TTC above $0.83$---and comfort (COM) softens only on the Extreme tier ($0.57$). The headline HD-Score is instead driven down primarily by route completion: $R_c$ falls from $0.84$ on Easy to $0.24$ on Extreme, and because it multiplies the aggregate (Eq.~\ref{eq:supp-cl-hdscore-agg}), it alone pulls the Extreme HD-Score to $0.19$ despite an HD-PDMS of $0.73$. Even the privileged teacher cannot make progress on many Extreme scenarios, consistent with the visually implausible initial configurations documented in Appendix~\ref{sec:supp-qualitative}.

\paragraph{Behavioral realism.}
Beyond driving quality, we ask whether the teacher's motion is realistic. Table~\ref{tab:supp-wosac-baseline} reports WOSAC likelihood metrics for the teacher's own rollouts in exported nuScenes traffic, under two control settings (ego-only with replayed traffic, and log-initialized multi-agent). Realism is high overall ($0.89$ and $0.86$), carried by near-perfect interaction and map-compliance likelihoods; the weakest axis is kinematic likelihood ($0.48$ and $0.51$), indicating that the teacher's speed and acceleration profile is the least human-like aspect of its otherwise safe, on-road behavior.

\begin{table}[htbp]
    \centering
    \caption{WOSAC metrics on exported nuScenes scenes. Evaluated across 18 exported nuScenes maps with 32 sampled policy rollouts per map. In the ego-with-replayed-traffic setting, the policy controls only the ego/SDC while surrounding vehicles replay the logs. In the log-initialized multi-agent setting, the policy controls valid vehicle tracks initialized from the logs and remaining logged entities provide replay context. Random-spawned agents are not reported in WOSAC because they do not have matching logged reference trajectories.}
    \label{tab:supp-wosac-baseline}
    \scriptsize
    \renewcommand{\arraystretch}{1.18}
    \setlength{\tabcolsep}{3pt}
    \begin{tabular}{@{}lrrrrrrr@{}}
    \toprule
    & \multicolumn{4}{c}{Realism components} & \multicolumn{2}{c}{Displacement} & \multicolumn{1}{c}{Spread} \\
    \cmidrule(lr){2-5}\cmidrule(lr){6-7}\cmidrule(l){8-8}
    Setting & Realism & Kin. & Inter. & Map & ADE & minADE & Std. \\
    \midrule
    \textbf{Ego with replayed traffic} & 0.894 & 0.476 & 1.000 & 0.996 & 5.313 & 2.745 & 0.027 \\
    \textbf{Log-initialized multi-agent} & 0.862 & 0.509 & 0.889 & 0.967 & 5.821 & 3.415 & 0.059 \\
    \bottomrule
    \end{tabular}
\end{table}

\noindent\textbf{WOSAC metric descriptions.}
\emph{Realism} is the WOSAC weighted meta-score over likelihood metrics.
\emph{Kin.} averages the kinematic likelihood metrics for linear and angular speed and acceleration.
\emph{Inter.} averages interaction likelihood metrics for collision, distance to nearest object, and time-to-collision.
\emph{Map} averages map-based likelihood metrics for distance to road edge and offroad indication.
\emph{ADE} is average displacement error, and \emph{minADE} is the best displacement error over stochastic rollouts.
\emph{Std.} is the standard deviation of per-scenario realism scores.

\section{Loss Ablation}
\label{sec:supp-loss-ablation}

\subsection{Loss Type Ablation}

\paragraph{Loss decomposition.}
We ablate the two terms of the alignment objective $\mathcal{L}_{\text{align}}=\mathcal{L}_{\text{act}}+\lambda\,\mathcal{L}_{\text{struct}}$ (\S\ref{sec:method:alignment}) by training the student with each term in isolation and with both, holding the rest of the recipe fixed, and reporting closed-loop HD-Score on the HUGSim scenarios together with the two component losses at convergence (Table~\ref{tab:supp-loss-decomposition}). Expected reading: structural-only loses action consistency, action-only loses feature stability, and the combined objective dominates both. We found performance to be largely insensitive to $\lambda$ and set it to $0.5$ by default.

\begin{table}[htbp]
    \centering
    \caption{ Loss-decomposition ablation. Closed-loop HD-Score and converged component losses for the action term, the structural term, and the combined objective.}
    \label{tab:supp-loss-decomposition}
    \begin{tabular}{@{}lrrr@{}}
    \toprule
    Objective & $\mathcal{L}_{\text{act}}\!\downarrow$ & $\mathcal{L}_{\text{struct}}\!\downarrow$ & HD-Score\,$\uparrow$ \\
    \midrule
    Action loss only ($\lambda{=}0$)                       & \textbf{0.0062} & 0.387 & 0.319 \\
    Structural loss only ($\mathcal{L}_{\text{act}}$ off)  & 0.412 & \textbf{0.010} & 0.307 \\
    Both (full, headline)                                  & 0.0312 & 0.121 & \textbf{0.490} \\
    \bottomrule
    \end{tabular}
\end{table}

\subsection{Structural Loss Rank Ablation}

We ablate the truncation rank of the structural alignment loss, comparing the low-rank-subspace similarity loss used in the headline (\S\ref{sec:method:alignment}; $k_p$ and $k_r$ at the $80\%$ cumulative-energy cutoffs) against intermediate ranks and a full-coordinate baseline that matches the pairwise similarity matrix over all feature coordinates, all else equal. The expected reading: full-coordinate matching matches or degrades closed-loop HD-Score, because the teacher features are sharply low-rank and redundant (\S\ref{sec:exp-rank}) so constraining every direction over-constrains the student on image-unrecoverable ones. This is the experiment that justifies the low-rank-subspace loss named in \S\ref{sec:method:alignment}.

\begin{table}[htbp]
    \centering
    \caption{Structural-loss rank ablation. Closed-loop HD-Score as the projection rank ranges from the headline low-rank subspace to the full coordinate space.}
    \label{tab:supp-struct-rank}
    \begin{tabular}{@{}lr@{}}
    \toprule
    Structural target & HD-Score\,$\uparrow$ \\
    \midrule
    Lower rank ($k_p{=}7,\,k_r{=}5$)                   & 0.417 \\
    Low-rank subspace ($k_p{=}13,\,k_r{=}9$, headline) & \textbf{0.490} \\
    Higher rank ($k_p{=}26,\,k_r{=}18$)                & 0.484 \\
    Full coordinate space (all directions)             & 0.444 \\
    \bottomrule
    \end{tabular}
\end{table}



\subsection{Low-Rank Subspace Diagnostics}

The structural loss does not align features in a fixed, precomputed subspace: at each step it recomputes a low-rank basis from the current batch (batch-wise centering and truncation to a per-stream rank $k_*$, standing for the partner rank $k_p$ or road rank $k_r$ of \S\ref{sec:method:alignment}; the diagnostics below are run per stream). The alignment target therefore depends on two knobs whose effect is not obvious a priori. The batch size $B$ controls \emph{which} axes the basis captures---as $B$ grows it is pulled toward large-scale, inter-cluster variance, while a small $B$ risks overfitting to sample-level noise---and the rank $k_*$ controls \emph{how much} of the teacher geometry survives truncation. We therefore verify that this batch-wise subspace is a faithful, useful target rather than an arbitrary one, along two axes: \emph{modal coverage} (does the batch-derived basis recover the dominant axes of the \emph{full} teacher distribution, a question about $B$?) and \emph{distinguishability} (does the truncated subspace still separate scenes that differ by small geometric perturbations such as ego or partner shifts and heading rotations, a question about $k_*$?).

\begin{figure}[htbp]
    \centering
    \includegraphics[width=\linewidth]{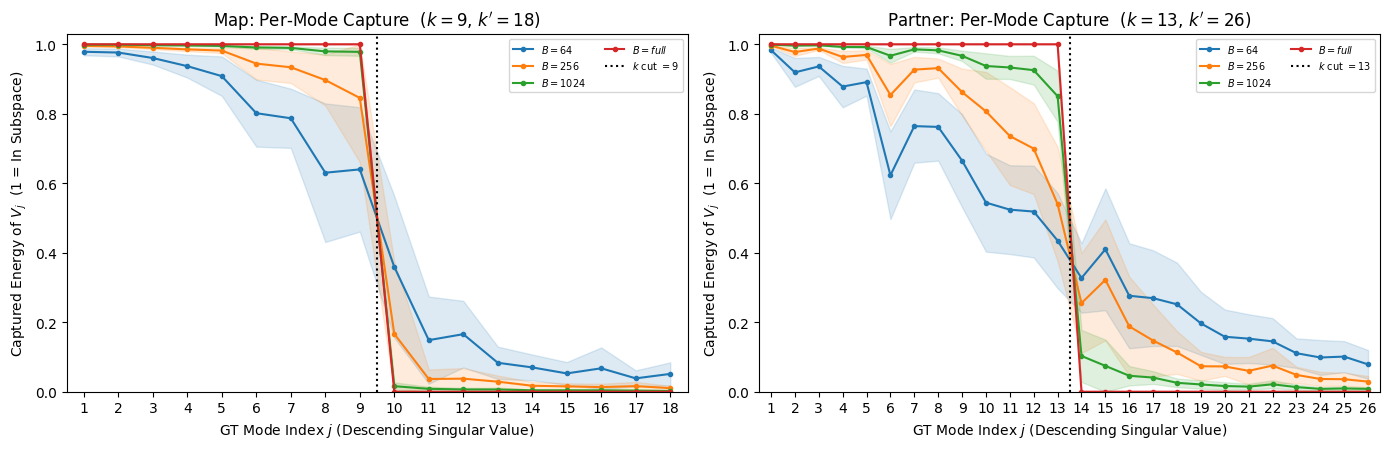}
    \caption{Modal coverage: ensemble-averaged squared projection of the batch-derived top-$k_*$ axes onto the reference top-$2k_*$ axes. At $B=1024$ the batch basis reliably captures the dominant teacher modes; coverage drops past the rank cutoff.}
    \label{fig:mode_capture}
\end{figure}

\paragraph{Modal coverage.}
We build the reference basis from the top singular vectors (SVD) of ${\sim}50{,}000$ teacher scene latents, and define modal coverage as the ensemble-averaged squared projection of the batch-derived top-$k_*$ axes onto this reference's top-$2k_*$ axes: writing $\{v_i\}_{i=1}^{k_*}$ for the batch axes and $P_{2k_*}$ for the orthogonal projector onto the reference top-$2k_*$ subspace, it is $\mathbb{E}_{\text{batch}}\!\big[\tfrac{1}{k_*}\sum_{i}\lVert P_{2k_*}v_i\rVert^2\big]\in[0,1]$, with $1$ meaning the batch basis lies entirely in the dominant teacher subspace. We compare against the top-$2k_*$ (rather than top-$k_*$) reference axes to tolerate ordering jitter between singular axes of nearby magnitude, and sweep batch sizes $B$ and ranks $k_*$.
Figure~\ref{fig:mode_capture} shows that at $B=1024$ the batch basis stably anchors to the dominant teacher modes; coverage drops only past the effective rank---the rank beyond which the teacher singular spectrum carries little additional energy (the $80\%$ cumulative-energy criterion of \S\ref{sec:method:alignment})---confirming that the truncation is principled.

\begin{figure}[htbp]
    \centering
    \begin{subfigure}[t]{0.48\linewidth}
        \includegraphics[width=\linewidth]{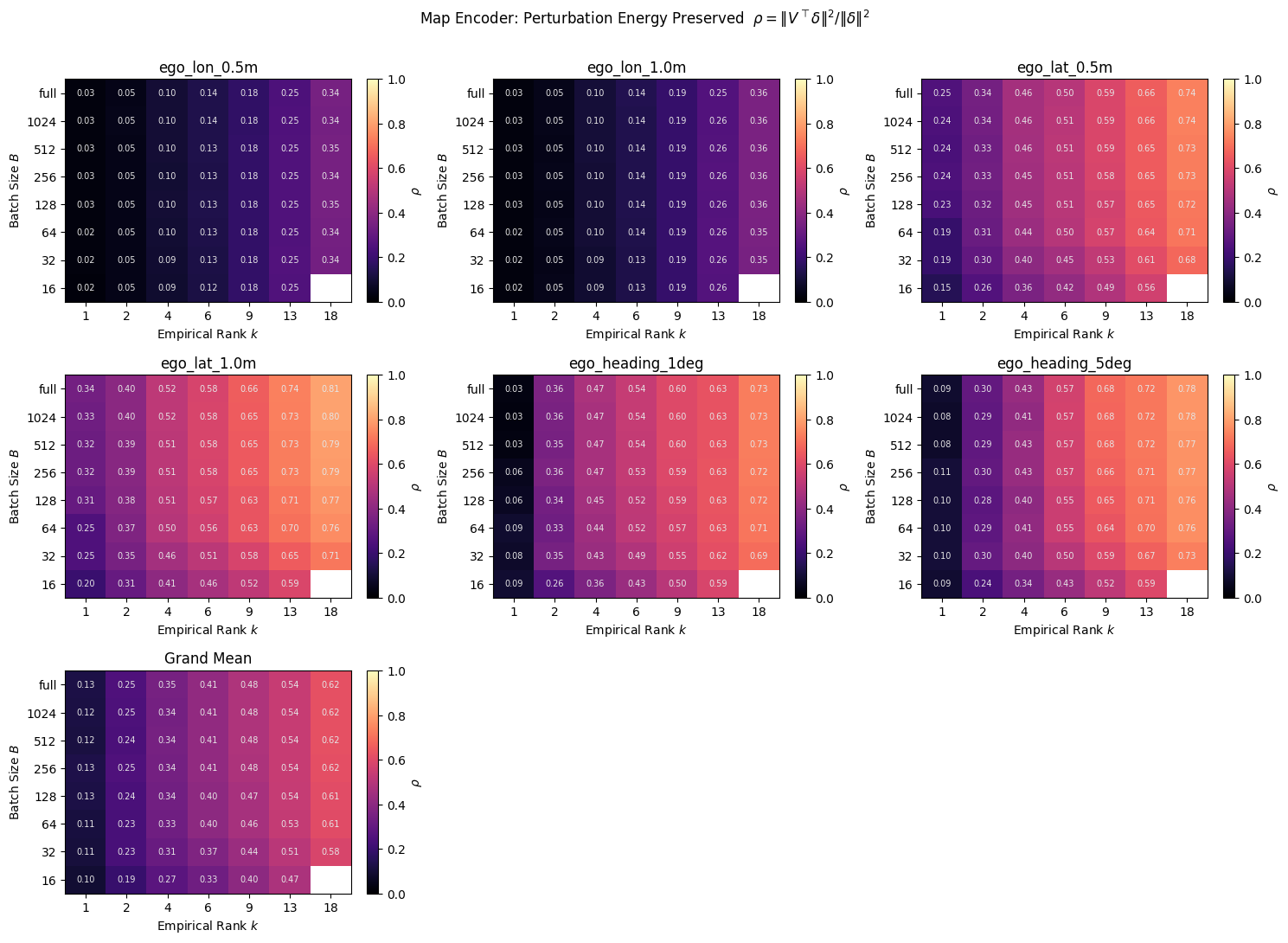}
        \caption{Map distinguishability. Partner augmentations omitted as the map embedding is invariant to partner locations.}
        \label{fig:map_energy}
    \end{subfigure}
    \hfill
    \begin{subfigure}[t]{0.48\linewidth}
        \includegraphics[width=\linewidth]{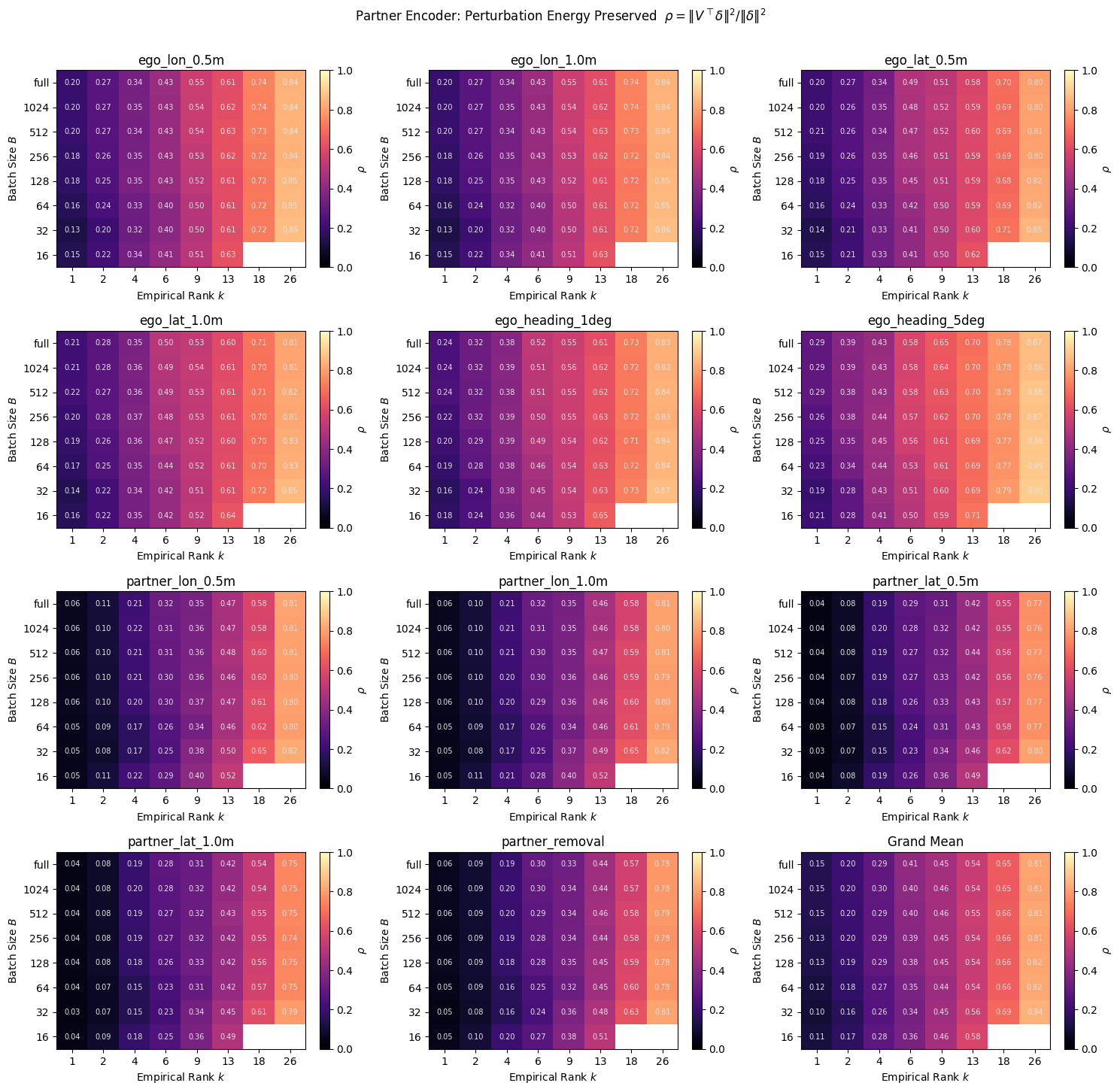}
        \caption{Partner distinguishability. Rank dependence mirrors the map results.}
        \label{fig:partner_energy}
    \end{subfigure}
    \caption{Fraction of perturbation energy preserved in the projected subspace as a function of batch size $B$ and rank $k_*$. Energy preservation depends primarily on rank and weakly on batch size.}
    \label{fig:perturbation_energy}
\end{figure}

\paragraph{Distinguishability.}
We perturb each scene by ego and nearest-partner longitudinal/lateral shifts and heading rotations, and measure what fraction of each perturbation's energy survives the low-rank projection: for a latent perturbation $\delta$ (the shift the augmentation induces in the teacher latent) and the batch projector $P_{k_*}$ onto the top-$k_*$ subspace, this is $\lVert P_{k_*}\delta\rVert^2/\lVert\delta\rVert^2\in[0,1]$, at most $1$ since $P_{k_*}$ is an orthogonal projection.
Figure~\ref{fig:perturbation_energy} shows that energy preservation depends strongly on $k_*$ and weakly on $B$: rank is the primary control over what geometry the alignment enforces.
Longitudinal ego shifts consistently show the lowest preservation, as driving scenes are approximately longitudinally invariant and this direction carries little variance in the teacher distribution.
Directions suppressed by the projection are not permanently unavailable to the student --- they can still be recovered via the action loss; the structural alignment only establishes a lower bound on what geometry is explicitly enforced.

\paragraph{Takeaway.}
The two diagnostics jointly justify the headline configuration: $B=1024$ yields a batch basis that faithfully tracks the full-distribution modes (modal coverage), and the $80\%$ cumulative-energy ranks used in \S\ref{sec:method:alignment} ($k_p{=}13,\,k_r{=}9$ in Table~\ref{tab:supp-struct-rank}) sit before the cutoff at which coverage drops, while still preserving the geometric separation that distinguishability tracks. The batch-wise low-rank target is thus a principled choice rather than an arbitrary one.

\section{Vision Backbone Ablation}
\label{sec:supp-backbone-ablation}

We compare the selected DINOv3 backbone against alternative image encoders (DINOv2 and ResNet-50) under an identical alignment recipe, reporting closed-loop HD-Score together with the structural and action losses at convergence (Table~\ref{tab:supp-backbone}).

\begin{table}[htbp]
    \centering
    \caption{Vision-backbone ablation. Closed-loop HD-Score and converged alignment losses under a fixed recipe for each image backbone. }
    \label{tab:supp-backbone}
    \begin{tabular}{@{}lrrr@{}}
    \toprule
    Backbone & $\mathcal{L}_{\text{act}}\!\downarrow$ & $\mathcal{L}_{\text{struct}}\!\downarrow$ & HD-Score\,$\uparrow$ \\
    \midrule
    DINOv3    & \textbf{0.0312} & \textbf{0.121} & \textbf{0.490} \\
    DINOv2    & 0.0519 & 0.197 & 0.488 \\
    ResNet-50 & 0.0991 & 0.221 & 0.459 \\
    \bottomrule
    \end{tabular}
\end{table}

\section{Alignment Decoupled from Demonstration Quality}
\label{sec:supp-demo-quality}

Since our standard training dataset is built upon expert demonstrations, it is natural to question whether strictly expert trajectories are required for the model to converge. In this section, we demonstrate that our method converges robustly even when trained on observation pairs collected from non-expert trajectories (Table~\ref{tab:supp-decoupled-demo}). As outlined in \S\ref{sec:method:alignment}, the alignment loss is evaluated against the teacher's action distribution for a given reconstructed scene state, independent of the policy that actually visited that state. Consequently, the quality of the rollout policy only alters the \emph{distribution of states} the student observes during alignment; it does not corrupt the supervision target at those states. 

\begin{figure}[h]
\centering
\includegraphics[width=\linewidth]{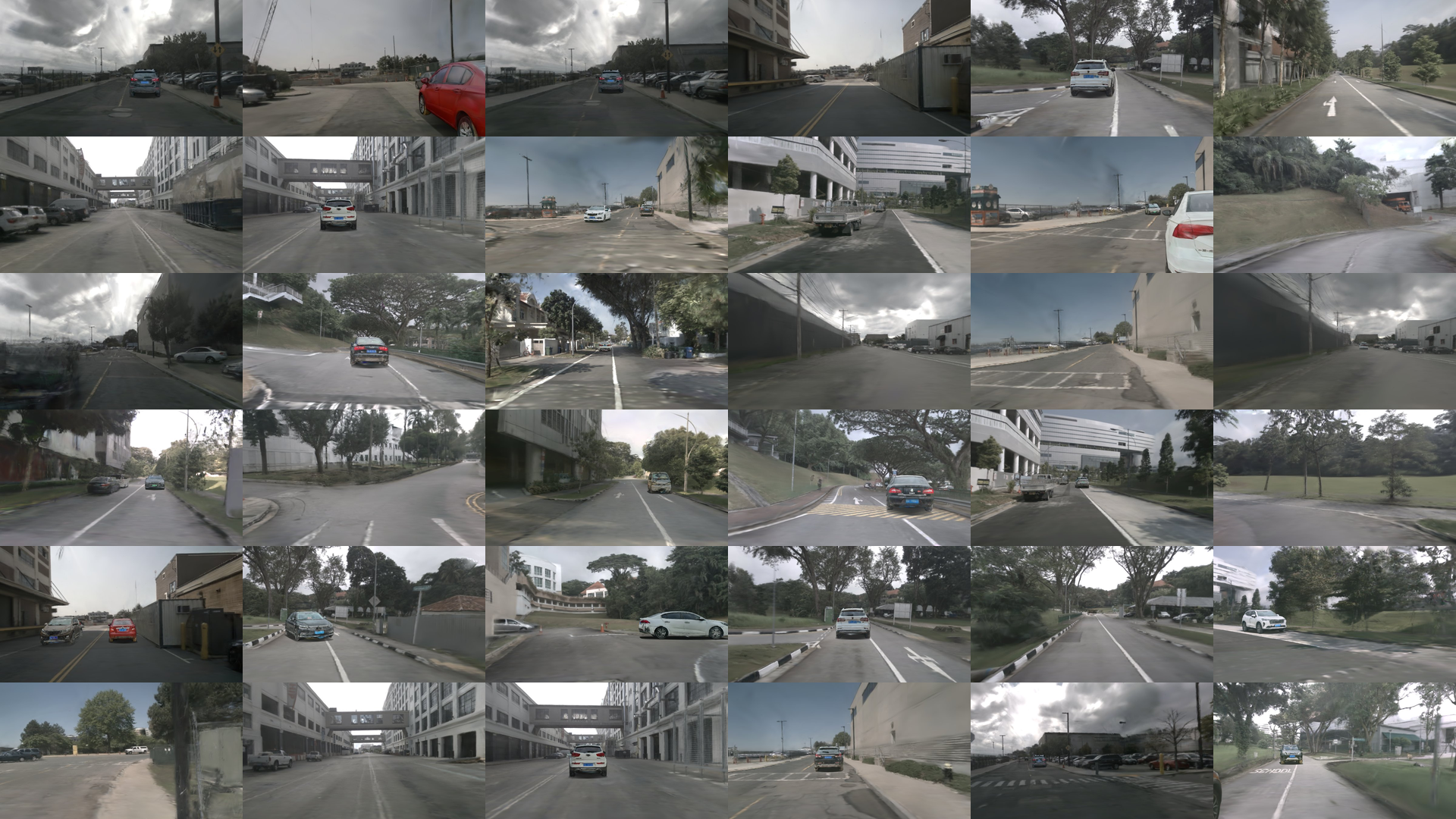}
\caption{\textbf{Front-camera samples from decoupled alignment data.} A grid of CAM\_F0 observations collected from HUGSim rollouts. These frames provide paired visual observations and reconstructed scene states for the alignment loss; their actions are not used as expert demonstrations, so rollout quality affects alignment only through the state coverage induced by the data-collection policy.}
\label{fig:demo-quality-front-samples}
\end{figure}

To empirically validate this, we collect an alignment dataset in the HUGSim simulator by randomly sampling actions; its size matches that of our full training dataset. Figure~\ref{fig:demo-quality-front-samples} shows randomly sampled front-camera observations from the resulting HUGSim collection.

\begin{table}[htbp]
    \centering
    \caption{Alignment outcome under different data-collection policies. Demonstration-collected and randomly sampled data yield broadly similar HD-Score ($0.490$ \vs $0.445$), indicating that alignment quality is gated mainly by state coverage rather than by the quality of actions taken during collection.}
    \label{tab:supp-decoupled-demo}
    \begin{tabular}{@{}lrrr@{}}
    \toprule
    Collection policy & $\mathcal{L}_{\text{act}}\!$ & $\mathcal{L}_{\text{struct}}\!$ & HD-Score\, \\
    \midrule
    Demonstration    & 0.0312 & \textbf{0.121} & \textbf{0.490} \\
    Random Sampled    & \textbf{0.0216} & 0.205 & 0.445 \\
    \bottomrule
    \end{tabular}
\end{table}

\clearpage

\section{Qualitative Results}
\label{sec:supp-qualitative}

\begin{figure}[htbp]
\centering
\includegraphics[page=1,width=\linewidth]{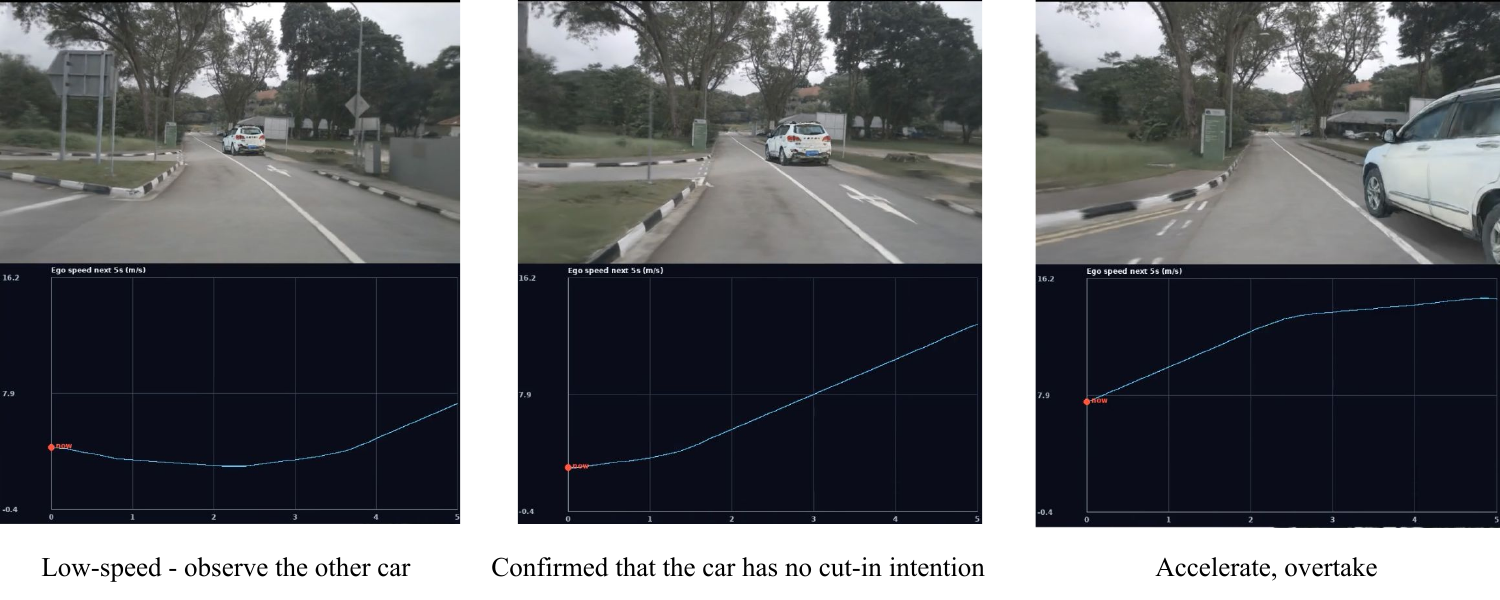}
\caption{\textbf{Cut-in negotiation.} The ego initially travels at low speed while monitoring a vehicle that could cut in. Once it confirms the vehicle has no cut-in intention (center), the ego accelerates and overtakes smoothly (right). The speed profile shows a brief deceleration followed by a sustained ramp-up, demonstrating that the policy withholds commitment until the other agent's intent is resolved.}
\label{fig:supp-qualitative-cutin}
\end{figure}

\begin{figure}[htbp]
\centering
\includegraphics[page=2,width=\linewidth]{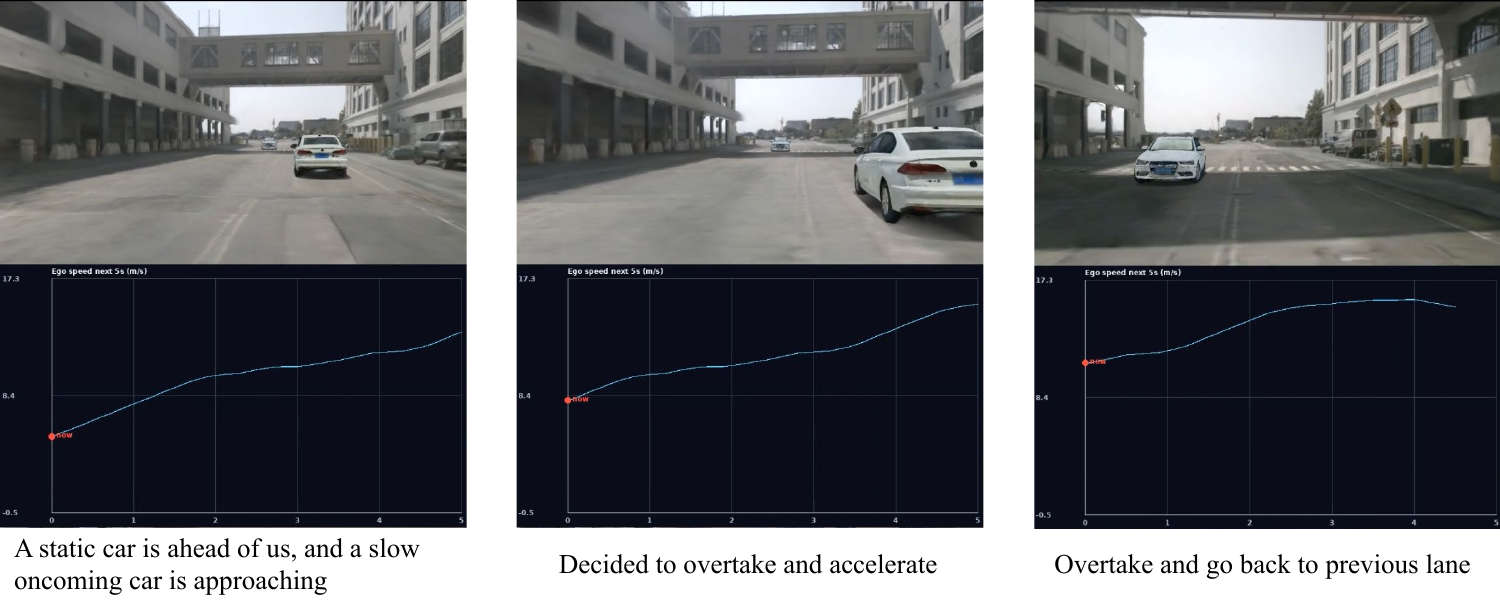}
\caption{\textbf{Overtaking with oncoming traffic.} A static vehicle blocks the lane while a slow oncoming car approaches. The policy decides to overtake (center), accelerates past the static vehicle, and merges back into the original lane before the oncoming car arrives (right). The speed profile rises throughout the maneuver, reflecting a committed overtake decision made under time pressure.}
\label{fig:supp-qualitative-overtake}
\end{figure}

\begin{figure}[htbp]
\centering
\includegraphics[page=3,width=\linewidth]{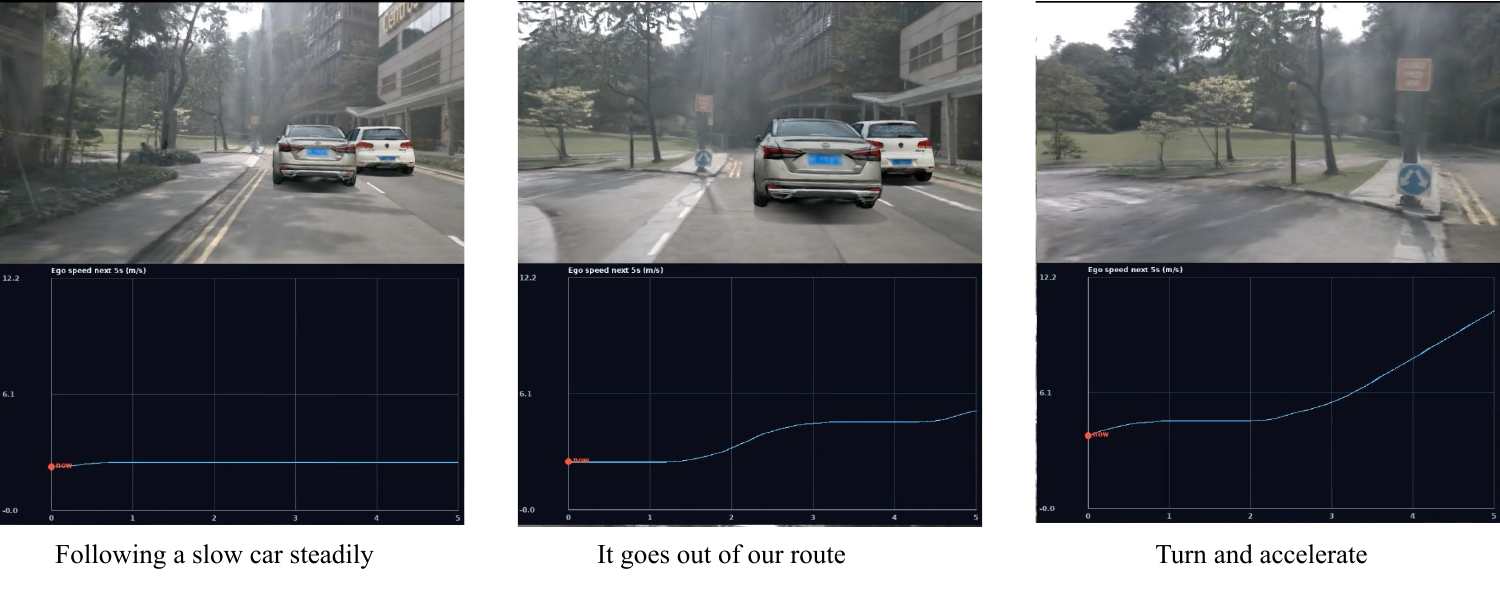}
\caption{\textbf{Lead vehicle yielding route.} The policy follows a slow lead vehicle at a steady speed (left). When the lead vehicle diverges from the ego's route (center), the ego turns onto its own path and accelerates (right). The speed profile is flat during the follow phase and then climbs sharply after the turn, showing the policy correctly disengages from the lead once it is no longer relevant.}
\label{fig:supp-qualitative-leadyield}
\end{figure}

\begin{figure}[htbp]
\centering
\includegraphics[page=4,width=\linewidth]{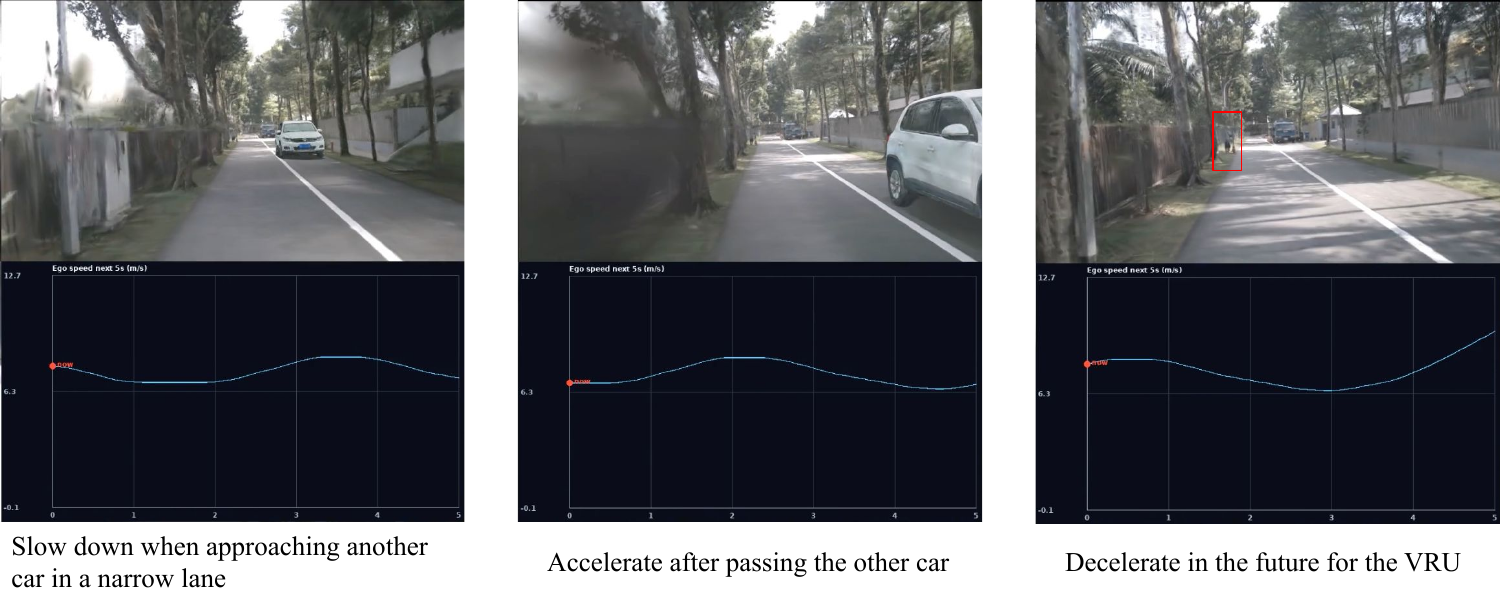}
\caption{\textbf{Narrow-lane passage.} In a narrow lane, the policy slows when approaching an oncoming vehicle (left), holds a reduced speed while passing (center), and then begins decelerating again in anticipation of a vulnerable road user detected ahead (right, red box). The speed profile oscillates rather than recovering fully, reflecting the policy's forward-looking awareness of the downstream hazard.}
\label{fig:supp-qualitative-narrowlane}
\end{figure}

\begin{figure}[htbp]
\centering
\includegraphics[page=5,width=\linewidth,trim={0pt 100pt 0pt 0pt},clip]{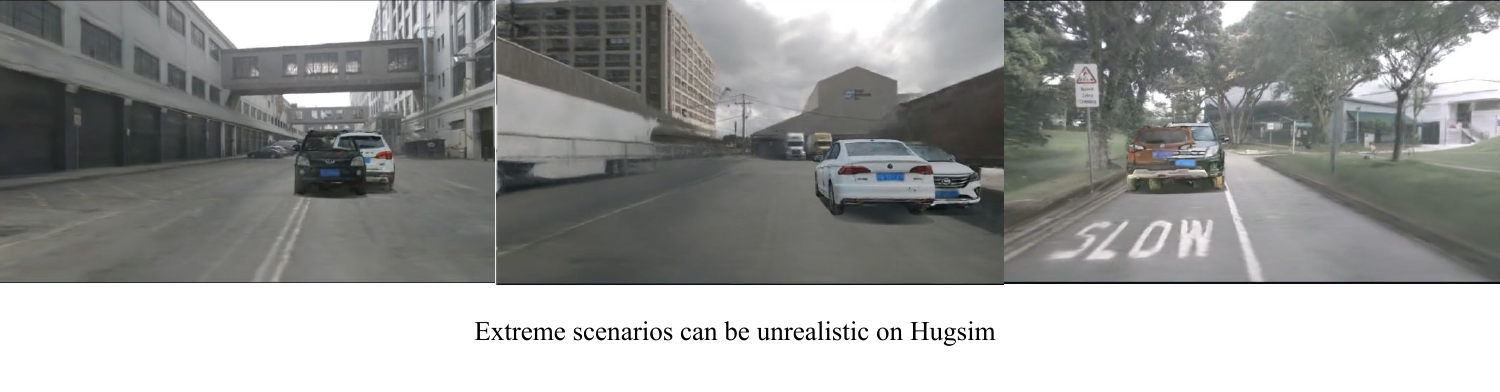}
\caption{\textbf{Visually implausible extreme-tier scenarios.} Extreme scenarios in the HUGSim benchmark can be visually unrealistic, with implausible initial configurations, or an occluded environment where other cars can pass through each other.}
\label{fig:supp-qualitative-extreme}
\end{figure}

We present closed-loop rollout scenarios drawn from the HUGSim evaluation split that illustrate the policy's behavior across a range of interactive driving situations. Each scenario shows three keyframes from the front-facing camera alongside the predicted ego speed profile for the next 5\,s, with the current speed marked in red. The four cases cover intent-uncertain cut-in negotiation (Figure~\ref{fig:supp-qualitative-cutin}), overtaking under oncoming-traffic time pressure (Figure~\ref{fig:supp-qualitative-overtake}), route-aware disengagement from a diverging lead vehicle (Figure~\ref{fig:supp-qualitative-leadyield}), and a narrow-lane passage with anticipatory deceleration (Figure~\ref{fig:supp-qualitative-narrowlane}). We additionally illustrate the visual implausibility of extreme-tier HUGSim scenarios (Figure~\ref{fig:supp-qualitative-extreme}), which imposes a ceiling on achievable HD-Score independent of policy quality.

\end{document}